%% file: iclr2027_conference.tex
\documentclass{article} 
\usepackage{iclr2027_conference,times}
\iclrfinalcopy 

\input{math_commands.tex}

\usepackage{hyperref}
\usepackage{url}
\usepackage{tikz}

\usepackage[utf8]{inputenc} 
\usepackage[T1]{fontenc}    
\usepackage{hyperref}       
\usepackage{url}            
\usepackage{booktabs}       
\usepackage{amsfonts}       
\usepackage{nicefrac}       
\usepackage{microtype}      
\usepackage{xcolor}         
\usepackage{wrapfig}

\usepackage{graphicx}
\usepackage{amsmath}
\usepackage{amssymb}
\usepackage{arydshln} 
\usepackage{amsfonts}
\usepackage{epsfig}
\usepackage{algorithm}
\usepackage{algpseudocode}
\usepackage{booktabs}
\usepackage{mathrsfs}

\usepackage{color}
\usepackage{textcomp}
\usepackage{bbding}
\usepackage{pifont}
\usepackage{wasysym}
\usepackage{subcaption}
\usepackage{xcolor,colortbl}
\usepackage{overpic}
\usepackage{microtype}
\usepackage{multirow}
\usepackage{makecell}
\usepackage{hhline}
\usepackage[american]{babel}
\usepackage{cuted}
\usepackage{ragged2e}
\definecolor{lightgray}{gray}{.94}
\definecolor{tinygray}{gray}{.96}

\newcommand{\ie}{\textit{i}.\textit{e}.}
\newcommand{\eg}{\textit{e}.\textit{g}.}

\usepackage{multirow}
\usepackage{epsfig}
\usepackage{adjustbox}

\title{HiFiPix: Bringing \underline{Hi}gh-\underline{Fi}delity Real-World Super-Resolution Back to \underline{Pix}el Space}

\author{
    Wenjie Li$^{1,2,}$\thanks{Equal contribution.},
	Du Chen$^{2,}$\footnotemark[1],
	Hui Wang$^{2}$,
    Linlong Fan$^{2}$,
    Youwei Lv$^{2}$,
    Zeyi Chen$^{2,3}$,\\ \bfseries 
	Zehan Chen$^{2,3}$,
    Jinwei Chen$^{2}$,
    Heng Guo$^{1,}$\thanks{Corresponding authors.},
    Zhanyu Ma$^{1}$,
	and Qingnan Fan$^{2,}$\footnotemark[2]\\
	$^1$Beijing University of Posts and Telecommunications \ 
	$^2$vivo BlueImage Lab \  $^3$Zhejiang University\\
}

\begin{document}

\maketitle

\vspace{-1.0cm}
\begin{figure*}[ht]
    \centering
    \begin{overpic}[width=0.99\linewidth]{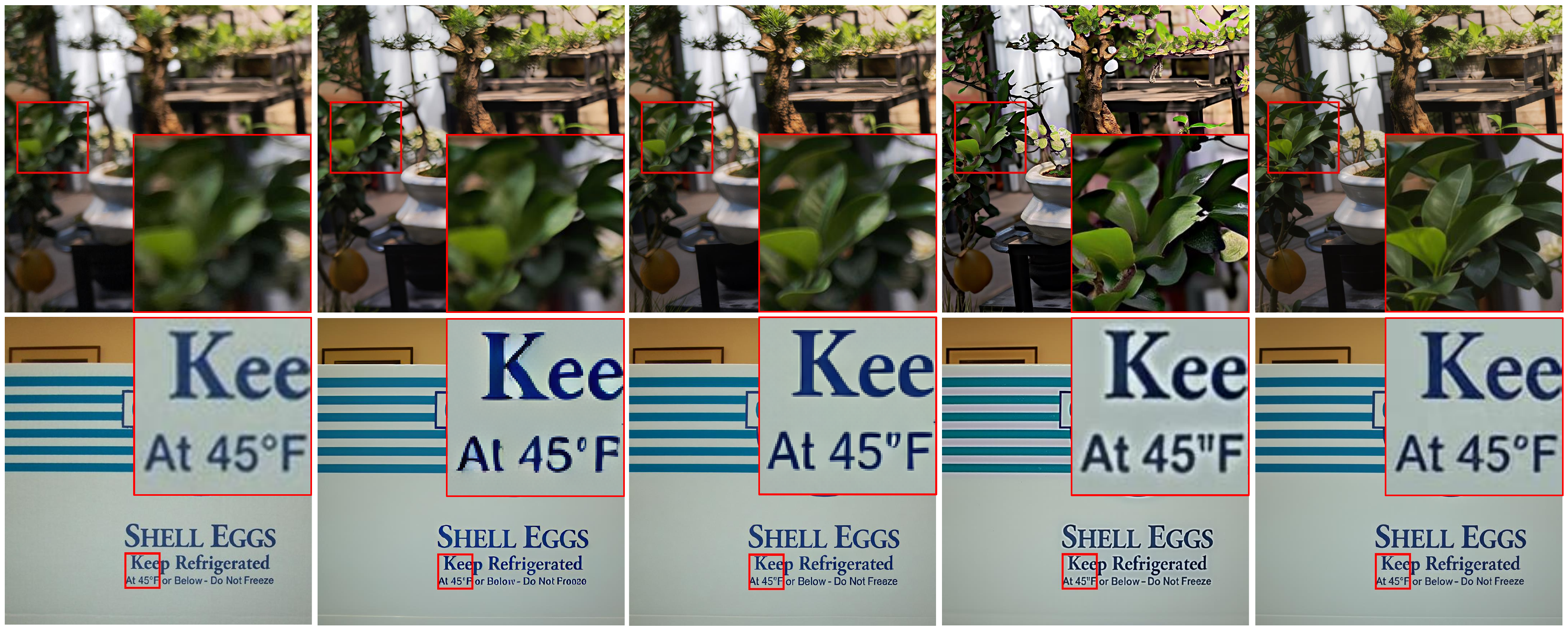}
    \put(3.3,-2){\color{black}{\fontsize{8.25pt}{1pt}\selectfont Real-World LQ}}
    \put(27.3,-2){\color{black}{\fontsize{8.25pt}{1pt}\selectfont TVT}}
    \put(46.0,-2){\color{black}{\fontsize{8.25pt}{1pt}\selectfont ODTSR}}
    \put(65.5,-2){\color{black}{\fontsize{8.25pt}{1pt}\selectfont LucidFlux}}
    \put(87.6,-2){\color{black}{\fontsize{8.25pt}{1pt}\selectfont \textbf{Ours}}}
    \end{overpic}
    \vspace{2.5mm}
    \caption{We propose HiFiPix, a Real-ISR framework built on a pixel-space large  generative model, achieving high-fidelity restoration of real-world LQ images and superior perceptual quality compared with existing latent-space Real-ISR methods.}
    \label{fig:abs}
\end{figure*}

\begin{abstract}
Diffusion-based real-world image super-resolution suffers from limited reconstruction fidelity due to information loss in VAE-based latent representations. Recently, pixel-space diffusion models offer a promising high‑faithfulness alternative by avoiding latent compression, but adapting them to Real-ISR remains challenging due to inherent patch boundary artifacts and degraded generative capability. To address the issues and maximize the potential of pretrained pixel-space diffusion models, we propose \textbf{HiFiPix}, a pixel‑space Real‑ISR framework capable of ultra high‑fidelity reconstruction and fine‑grained visual generation effects. To suppress patch boundary artifacts, we penalize unrealistic high-frequency periodic components with a Cross-Scale Wavelet Discriminator (CWSD) for image-level supervision, and Mixup Contrastive Self-Distillation (MixCSD) for intermediate representation regularization. Furthermore, we uncover a heterogeneous generation-fidelity trade-off across low-rank adapters (LoRA) and propose Selective LoRA Pruning (LoRA-SP) to adaptively prune LoRA layers, enhancing generative capability with minimal deterioration of reconstruction fidelity. Extensive experiments demonstrate that HiFiPix consistently outperforms existing state‑of‑the‑art latent‑space methods in terms of both restoration accuracy and perceptual quality by a large margin. Code and models will be released.

\end{abstract}

\section{Introduction}
Real‑world image super‑resolution (Real-ISR) targets the recovery of low‑quality (LQ) images suffering from complex unknown degradations, with the goal of reconstructing high‑quality (HQ) counterparts that exhibit both high fidelity and satisfying perceptual quality. Initially, GAN‑based algorithms \cite{wang2021real, chen2023human} employ a discriminator to match the distribution of reconstructed images with that of real‑scenario images. However, lacking powerful generative priors, GAN‑based models exhibit limited capability for recovering fine‑grained details.

Over the past few years, diffusion-based generative models~\cite{rombach2022high,flux-2-2025} have advanced Real-ISR by producing highly photo-realistic details. However, existing Real-ISR methods~\cite{dong2025tsd,fang2026one} typically rely on the Variational Autoencoder (VAE)~\cite{esser2021taming} to compress LQ inputs into the latent space for efficient denoising. This irreversible compression discards the subtle textures from LQ inputs, often leading to structural inconsistencies, ghosting artifacts, and ultimately compromised restoration fidelity. To mitigate this issue, prior works try to improve input quality~\cite{lin2024diffbir}, reduce the VAE compression ratio~\cite{yi2025fine}, or inject LQ features into denoising networks~\cite{yang2024pixel}. However, these strategies fail to fundamentally address the information loss introduced by latent space compression, yielding reduced generative capability (\eg, blurry leaves in the top row of Fig.~\ref{fig:abs}) or limited fidelity gains (\eg, distorted characters in the bottom row of Fig.~\ref{fig:abs}). As a result, restoring highly faithful textures while preserving excellent visual quality still remains a significant challenge in Real-ISR.

\begin{figure*}[t]
\begin{overpic}[width=0.99\linewidth]{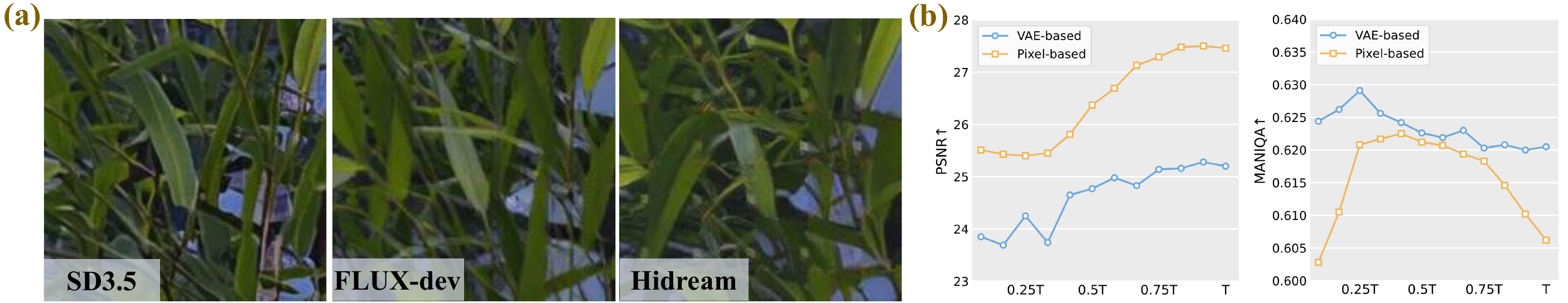}
\end{overpic}
\vspace{-1mm}
\caption{Fine-tuning pixel models faces two challenges compared with latent models: \textbf{(a)} without VAE decoders to preserve spatial continuity, pixel-space models are prone to patch boundary artifacts; \textbf{(b)} training progress tends to favor high-fidelity restoration at the expense of generative capability.}
\label{fig: existing_issues}
\vspace{-4mm}
\end{figure*}

More recently, pixel‑level text-to-image (T2I) diffusion models~\cite{yu2026pixeldit,hidreamolimage} have demonstrated effectiveness in mitigating reconstruction accuracy issues by removing the VAE component. However, roughly fine-tuning pixel-space models to Real-ISR via LoRA adapters still faces two key challenges. First, pixel-space models typically adopt patch-based tokenization, while different tokens only interact within pure MLP-based attention blocks. Unlike VAE-based ones~\cite{saharia2022image,esser2024scaling}, whose convolutional decoders naturally preserve spatial continuity, pixel-space decoding does not explicitly enforce information interaction across patch boundaries. This leads to discontinuities between patches and periodic grid artifacts, particularly in texture-rich regions. As shown in Fig.~\ref{fig: existing_issues} (a), under the same training strategy, pixel-space models suffer from more severe periodic artifacts than latent models~\cite{esser2024scaling,flux-2-2025}, degrading the structural consistency across neighboring patches. Second, as depicted in Fig.~\ref{fig: existing_issues} (b), although removing the VAE eliminates latent compression, it discards the compact semantic representations, making high-level semantic modeling more challenging in pure pixel space. Therefore, this weakens the generative prior and biases models toward deterministic reconstruction rather than plausible detail synthesis. Consequently, models tend to overfit reasonable but over-smoothed structures, while losing their ability to generate missing high-frequency details, especially under severe degradations.

To address these challenges, we propose \emph{HiFiPix}, a pixel-space framework for high-fidelity Real-ISR. We first focus on patch-boundary artifacts arising from pixel-space decoding and observe that they mainly manifest as discontinuities in high-frequency structures. This motivates us to exploit coherent high-frequency information for cross-patch continuity. Specifically, we introduce a \emph{Cross-Scale Wavelet Discriminator (CSWD)} that propagates high-frequency features across scales for boundary-aware image-level supervision, and further propose a \emph{Mixup Contrastive Self-Distillation (MixCSD)} that leverages an HQ-mixed branch to regularize intermediate representations at feature levels. Together, they provide complementary external and internal supervision to suppress patch-boundary artifacts. Furthermore, we address degradation of generative capability during training. We observe that LoRA updates across layers are not equally necessary: some deep layers merely echo transformations already made by shallower layers, contributing to overfitting pixel-level characteristics rather than preserving high-level semantics. We thus propose \emph{Selective LoRA Pruning (LoRA-SP)}, which detects and prunes such redundant layers via Centered Kernel Alignment-based representational~\cite{kornblith2019similarity}, protects fidelity-critical layers from pruning, and progressively freezes rest, thereby constraining fine-tuning capacity precisely where overfitting risks occur. 

Furthermore, existing Real-ISR benchmarks are hindered by low-quality ground truth, suffering from missing fine textures~\cite{cai2019toward,wei2020component} or background de-focus~\cite{wu2026vosr}. As noted in A-FINE~\cite{chen2025toward}, generative restoration models may yield images visually superior to these flawed GT references, rendering nearly all reference-based fidelity metrics unreliable. To address this issue, we construct \emph{Dual-Q-Bench}, an ultra-high-quality paired real-world ISR benchmark captured via screen photography, whose HQ reference images feature rich textures and minimal background de-focus. Dual-Q-Bench enables reliable evaluation of both fidelity and perceptual quality. Our contributions are summarized as follows:

\begin{itemize} 
\item We propose \emph{HiFiPix}, a Real-ISR method built on a pretrained pixel-level model, shifting Real-ISR from latent-space to pixel-space modeling, achieving high-fidelity restoration without sacrificing generative capability or inference efficiency.
\item We propose \emph{CSWD} that propagates high-frequency features across scales for boundary-aware image-level supervision, and \emph{MixCSD}, which distills clean representations to enhance patch-level discriminability, jointly mitigating patch-boundary artifacts.
\item We propose \emph{LoRA-SP}, which identifies functionally redundant LoRA layers via representational drift and progressively prunes them while protecting fidelity-critical layers, preserving the pretrained generative prior during supervised fine-tuning.
\item We introduce \emph{Dual-Q-Bench}, a high-quality paired Real-ISR benchmark that addresses the low-quality GT and background de-focus in existing benchmarks, enabling more reliable evaluation of both generative capability and reconstruction fidelity.
\end{itemize}

\section{Related Works}
\subsection{Real-world Image Super-Resolution}

With the advent of large-scale T2I models, diffusion-based Real-ISR methods have surpassed GAN-based approaches. Existing methods utilize LQ input to control diffusion models for faithful restoration by leveraging pluggable adapters~\cite{chen2025faithdiff}, dual-stream architectures~\cite{yang2024pixel}, ControlNet~\cite{yu2024scaling}, initialized restoration modules~\cite{lin2024diffbir}, semantic conditioning~\cite{wu2026vosr}, and large-scale high-quality training datasets~\cite{yu2024scaling}. Furthermore, to fully use input information, methods like OSEDiff~\cite{wu2024one} and TSD-SR~\cite{dong2025tsd} replace random noise with LQ images, while S3Diff~\cite{zhang2024degradation} and ODTSR~\cite{fang2026one} incorporate GAN loss to enhance perceptual quality. TVT~\cite{yi2025fine} increases the VAE latent resolution to improve fidelity, whereas LucidFlux~\cite{feilucidflux} and StrSR~\cite{wang2026spectral} improve fidelity by fine-tuning FLUX~\cite{flux-2-2025} or Z-image~\cite{cai2025z} with increased VAE channel capacity. Nevertheless, these methods still operate in latent spaces, where VAE compression inevitably limits reconstruction fidelity. In contrast, we perform Real-ISR directly in pre-trained pixel spaces~\cite{hidreamolimage}, enabling high-fidelity and perceptually pleasing Real-ISR.

\subsection{Pixel-space Diffusion Models}
Early pixel-space diffusion models~\cite{dhariwal2021diffusion,saharia2022image} suffered from prohibitive computational costs at high resolutions. Recent works~\cite{chen2025pixelflow,wang2025pixnerd,li2026back,ma2026pixelgen,lu2026one,ma2026deco,yu2026pixeldit} have revisited pixel-space diffusion by improving its training efficiency, sampling strategy, and architectural design. Among them, PixelFlow~\cite{chen2025pixelflow} enables efficient end-to-end pixel-space generation, while JiT~\cite{li2026back} and PixelDiT~\cite{yu2026pixeldit} further improve the scalability and efficiency of pixel-space modeling. HiDream-O1-Image~\cite{hidreamolimage} extends this paradigm to a large-scale pixel-space T2I diffusion transformer with a unified generative architecture. Inspired by these advances, we explore fine-tuning pre-trained pixel-space T2I models for Real-ISR, aiming to achieve high-fidelity restoration and visually pleasant generation without VAE compression.

\section{Method}


\subsection{Empirical Observations}\label{sec:Insight}

\begin{figure*}[t]
 \centering
  \begin{overpic}[width=0.9\linewidth]{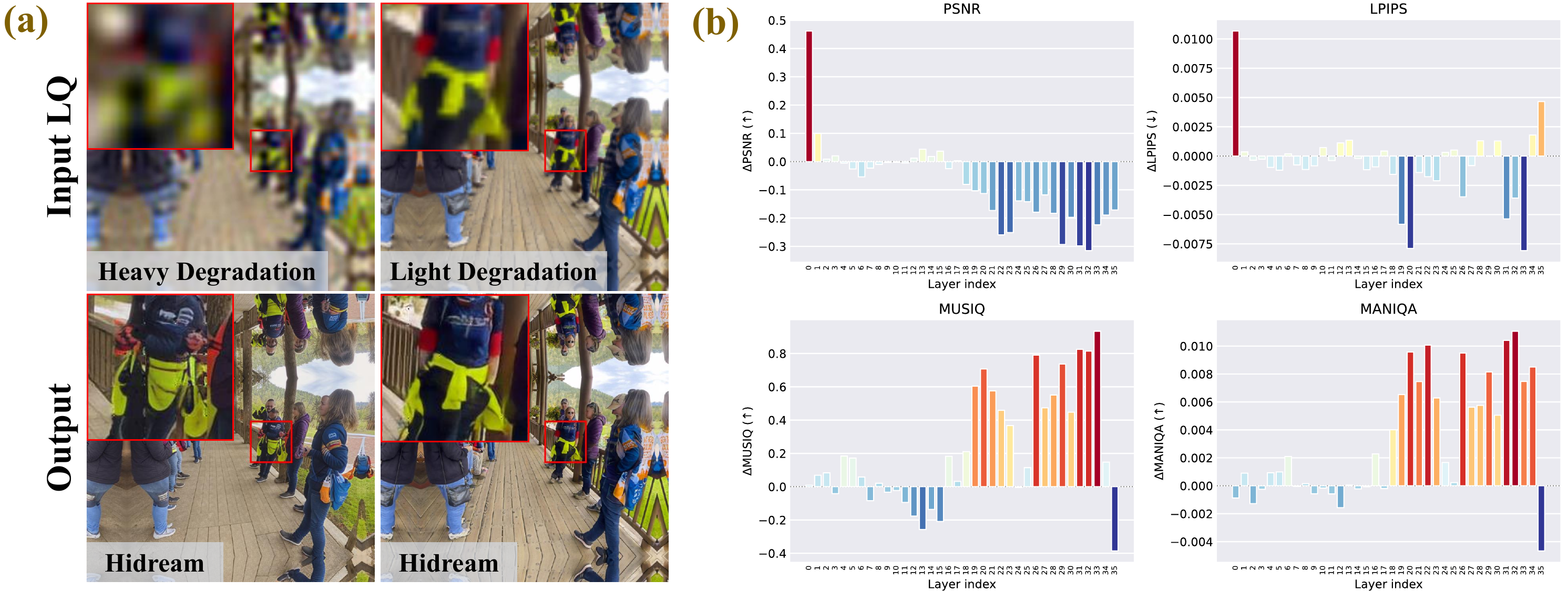}
	\end{overpic}
  \vspace{-1.5mm}
  \caption{\small{\textbf{(a)} Heavier boundary artifacts when super-resolving extremely blurry input. \textbf{(b)} Removal of individual LoRA layers reveals trade-offs between reconstruction and generative capabilities.}}
   \label{fig: insight}
\vspace{-4mm}
\end{figure*}

Following classical one-step fine-tuning paradigms~\cite{yi2025fine}, we identify two critical insights when applying HiDream-O1-Image for Real-ISR. \textbf{First}, patch-boundary artifacts may not arise solely from the patch-wise architecture of pixel-space models. As shown in Fig.~\ref{fig: insight} (a), grid-like artifacts become more pronounced under severe degradation than under mild degradation, suggesting that imperfect high-frequency reconstruction is an important cause of boundary artifacts, as severe degradation makes high-frequency reconstruction more challenging. This observation motivates a low-cost and architecture-agnostic strategy to explicitly regularize high-frequency components. Specifically, we design the Cross-Scale Wavelet Discriminator, which penalizes spurious boundary artifacts via multi-scale information constraints. Besides, we propose Mixup Contrastive Self-Distillation to ensure faithful generation of genuine high-frequency details.


\textbf{Second}, simply applying LoRA layers yields SR outputs lacking fine-grained details, indicating that such a general strategy drastically reduces the generative capability of pixel-space models. As shown in Fig.~\ref{fig: insight} (b), we perform ablation studies to investigate individual LoRA layers, then reveals that generative capacity and fidelity are non-uniformly entangled across depth: removing certain layers boosts generative expressiveness at negligible fidelity cost, whereas removing others yields a similar generative gain but at the expense of severe fidelity degradation. This heterogeneity rules out uniform layer pruning or fine-tuning as an effective strategy; therefore, we propose Selective LoRA Pruning that identifies and removes layers that mainly reduce the generation ability.

\subsection{Cross-Scale Wavelet Discriminator}
\label{sec:cswd}
\begin{figure*}[t]
	\begin{overpic}[width=0.99\linewidth]{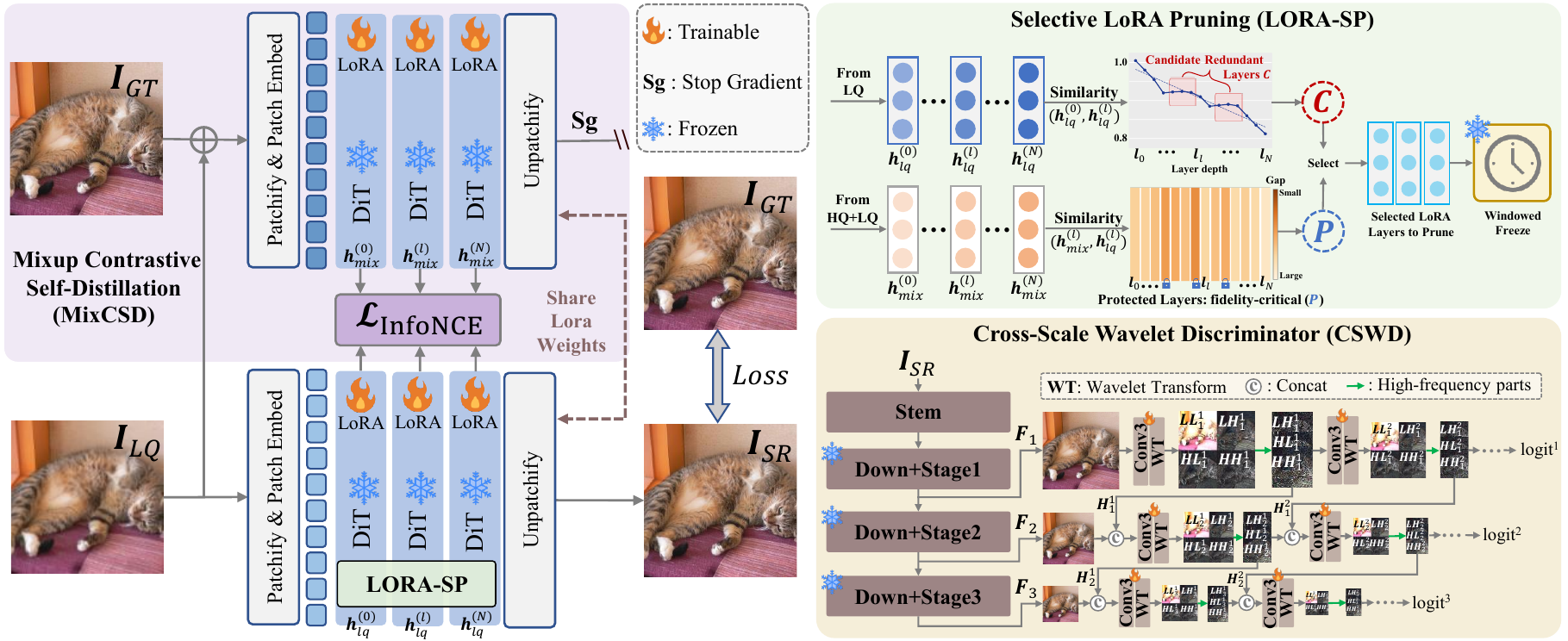}
	\end{overpic}
	\vspace{-2mm}
    \caption{Architecture of HiFiPix. We introduce MixCSD and CSWD to alleviate patch boundary artifacts, and propose LoRA-SP to enhance the model’s generative capability.}
	\label{fig: main}
	\vspace{-3mm}
\end{figure*}

Inspired by ~\cite{wu2025omgsr,wang2026spectral} that leverage DINOv3-ConvNeXt~\cite{simeoni2025dinov3} as a discriminator to enhance the visual quality, we propose \textbf{Cross-Scale Wavelet Discriminator (CSWD)} to extract multi-scale features and penalize the unrealistic high-frequency periodic components from boundaries, thereby alleviating patch boundary artifacts. As shown in Fig.~\ref{fig: main}, given the super-resolved image $I_{\mathrm{SR}}$, the frozen DINOv3-ConvNeXt first extracts multi-level features \hbox{$\boldsymbol{F}_{1}$~$\in$~$\mathbb{R}^{\frac{H}{4}\times\frac{W}{4}\times C}$}, \hbox{$\boldsymbol{F}_{2}$~$\in$~$\mathbb{R}^{\frac{H}{8}\times\frac{W}{8}\times 2C}$}, and \hbox{$\boldsymbol{F}_{3}$~$\in$~$\mathbb{R}^{\frac{H}{16}\times\frac{W}{16}\times 4C}$}. Different from~\cite{wu2025omgsr} that relies on pooling or convolution layers for multi-scale down-sampling, we employ the wavelet transform (WT) to decompose features into a low-frequency sub-band $\boldsymbol{LL}$ and three high-frequency sub-bands $\left\lbrace \boldsymbol{LH},\boldsymbol{HL},\boldsymbol{HH}\right\rbrace$, all with half the spatial resolution of input features. We discard $\boldsymbol{LL}$ and concatenate the three high-frequency sub-bands along the channel dimension, thereby focusing the discriminator on edges, textures, and periodic artifacts. The concatenated features are then compressed by trainable convolutional layers, while cascaded WT-convolution layers progressively extract deeper high-frequency representations and produce a pixel-wise discriminative score map.

To further exploit cross-scale information, we introduce a cross-scale high-frequency injection mechanism. Specifically, high-frequency sub-bands $\left\lbrace\boldsymbol{LH}_i^k,\boldsymbol{HL}_i^k,\boldsymbol{HH}_i^k\right\rbrace$ extracted by the $k$-th WT block $WT_k$ from level $\boldsymbol{F}_i$ are aggregated via a $1\times1$ convolution, then injected into features of next level $\boldsymbol{F}_{i+1}$. This progressively propagates fine-grained high-frequency details from shallow to deep layers, enhancing the discriminator's ability to detect multi-scale periodic boundary artifacts.
\begin{equation}
WT_k\left(Conv\left(\boldsymbol{F}_i\right)\right)=\left\lbrace\boldsymbol{LL}_i^k,\boldsymbol{LH}_i^k,\boldsymbol{HL}_i^k,\boldsymbol{HH}_i^k\right\rbrace,
\boldsymbol{H}_{i}^k=Conv\left(\boldsymbol{LH}_{i}^k+\boldsymbol{HL}_{i}^k+\boldsymbol{HH}_{i}^k\right),
\end{equation}
\begin{equation}
WT_{k-1}\left(Conv\left(Concat\left(\boldsymbol{H}_{i}^k,\boldsymbol{F}_{i+1}\right)\right)\right)=\left\lbrace \boldsymbol{LL}_{i+1}^k,\boldsymbol{LH}_{i+1}^k,\boldsymbol{HL}_{i+1}^k,\boldsymbol{HH}_{i+1}^k\right\rbrace.
\end{equation}

The multi-level discriminator progressively produces pixel-wise logit maps $\boldsymbol{M} \in \lbrace\mathbb{R}^{\frac{H}{4}\times\frac{W}{4}\times 1},\allowbreak \mathbb{R}^{\frac{H}{8}\times\frac{W}{8}\times 1},\allowbreak \mathbb{R}^{\frac{H}{16}\times\frac{W}{16}\times 1}\rbrace$ for PatchGAN-style~\cite{isola2017image} local realism discrimination.


\subsection{Mixup Contrastive Self-Distillation}\label{sec:Distillation}
To further suppress boundary artifacts while promoting the faithful generation of clean details, as shown in Fig.~\ref{fig: main}, we introduce \textbf{Mixup Contrastive Self-Distillation (MixCSD)} at intermediate layers of DiT blocks. Specifically, for each training batch, we construct two forward branches sharing the same set of LoRA weights \hbox{$\boldsymbol{f}_\theta$} and the same DiT weights. The anchor branch takes the pure LQ image \hbox{$\boldsymbol{I}_{LQ}$} as input, with gradients flowing normally, and extracts patch-token hidden states at DiT layer $\ell$:
\begin{equation}
\boldsymbol{h}^{\ell}_{lq} = f_\theta^{(\ell)}(\boldsymbol{I}_{LQ}) \in \mathbb{R}^{B\times N\times D},
\end{equation}
where $B$, $N$, and $D$ denote the batch size, the number of patch tokens, and the hidden dimension, respectively. Here we omit the DiT weights in all equations for simplicity of expression. The positive branch takes a mixture of GT and LQ as input with mixing ratio $ \alpha\in\left\lbrack0,1\right\rbrack $, then feeds forward through sharing the anchor branch weights under a stop-gradient operator $\mathrm{sg}$:
\begin{equation}
\boldsymbol{h}^{\ell}_{mix} = \mathrm{sg}\!\left(f_\theta^{(\ell)}(\alpha\, \boldsymbol{I}_{GT} + (1-\alpha)\, \boldsymbol{I}_{LQ})\right),
\end{equation}
since the mixed input injects genuine high-frequency content and consistent cross-patch boundary structure, $\boldsymbol{h}^{\ell}_{mix}$ carries more coherent neighborhood features at patch boundaries than $\boldsymbol{h}^{\ell}_{lq}$, serving as an implicit target for the anchor branch along the "boundary-consistency" dimension. After $\ell_2$-normalization of the anchor (\eg, $\bar{\boldsymbol{h}}_{lq,i}= \frac{\boldsymbol{h}_{lq,i}}{\left\|\boldsymbol{h}_{lq,i}\right\|_2}$), we treat each patch-token as a contrastive unit (flattened index $i=1,...,BN$) and define an InfoNCE loss~\cite{oord2018representation} with temperature $\tau$:
\begin{equation}
\mathcal{L}_{\text{InfoNCE}}^{(\ell)} = -\frac{1}{BN}\sum_{i=1}^{BN}
\log \frac{\exp\!\left(\bar{\boldsymbol{h}}_{lq,i}^{\top}\bar{\boldsymbol{h}}_{mix,i}/\tau\right)}
{\displaystyle\sum_{j=1}^{BN}\exp\!\left(\bar{\boldsymbol{h}}_{lq,i}^{\top}\bar{\boldsymbol{h}}_{mix,j}/\tau\right)},
\end{equation}
where $j=i$ term in the denominator corresponds to the positive pair and the remaining $ BN-1 $ terms serve as negatives, and $\top$ denotes the matrix transpose. Negatives are sampled from other patches in the same image or other images in the same training batch. This encourages each anchor to approach its high-frequency-consistent target while moving away from other patches, particularly spatially adjacent ones, thereby suppressing patch boundary artifacts. A stop-gradient operation enforces one-way supervision from the positive branch to the anchor branch, preventing representational collapse. Multi-layer aggregation and step-gated weighting follow the same formulation as follows:
\begin{equation}
\mathcal{L}_{\text{InfoNCE}} = \lambda_{\text{InfoNCE}}(t)\cdot \frac{1}{|\mathcal{S}|}\sum_{\ell \in \mathcal{S}} \mathcal{L}_{\text{InfoNCE}}^{(\ell)},
\lambda_{\text{InfoNCE}}(t) = \begin{cases} \lambda_{\text{InfoNCE}}^{\max}, & t \ge t_{\text{start}} \\ 0, & t < t_{\text{start}} \end{cases},
\end{equation}
where $\mathcal{S}$ is the number of supervised DiT layers, $t$ is the training step. $\lambda_{\text{InfoNCE}}^{\max}$ and $t_{\text{start}}$ denote the target loss weight and the step at which it is activated, respectively.


\subsection{Selective LoRA Pruning}
\label{sec:slp}
As shown in Fig.~\ref{fig: main}, we introduce \textbf{Selective LoRA Pruning (LoRA-SP)}, a training-time mechanism that leverages the network's internal representations to identify and freeze functionally redundant LoRA branches, thereby improving generative capacity without sacrificing restoration fidelity. 


At each probing step, we forward the degraded input \(\boldsymbol{I}_{LQ}\) and mixture \(\boldsymbol{x}_{mix} = \alpha \boldsymbol{I}_{GT} + (1-\alpha) \boldsymbol{I}_{LQ}\) through the network to gather per‑layer hidden states \(h^{(l)}\). Layer‑wise similarity is quantified by linear Centered Kernel Alignment (CKA)~\cite{kornblith2019similarity} and defined as:

\begin{equation}
\mathrm{CKA}(\boldsymbol{H}_1, \boldsymbol{H}_2) = \frac{\lVert \boldsymbol{H}_1^{\top} \boldsymbol{H}_2 \rVert_F^2}{\lVert \boldsymbol{H}_1^{\top} \boldsymbol{H}_1\rVert_F \, \lVert \boldsymbol{H}_2^{\top} \boldsymbol{H}_2\rVert_F},
\end{equation}

where $\boldsymbol{H}_1$, $\boldsymbol{H}_2$ and $F$ denote two intermediate features and Frobenius norm, respectively. We derive two EMA‑smoothed per‑layer metrics: anchor similarity \(\bar{\boldsymbol{a}}^{(l)}\) via \(\mathrm{CKA}(\boldsymbol{h}^{(l)}_{lq}, \boldsymbol{h}^{(r)}_{lq})\) with shallow reference layer $r$ (expected to decay under generative processing), and fidelity gap \(\bar{\boldsymbol{g}}^{(l)} = 1-\mathrm{CKA}(\boldsymbol{h}^{(l)}_{lq}, \boldsymbol{h}^{(l)}_{mix})\), measuring layer‑$l$’s representation shift toward high‑quality targets. A layer becomes a redundancy candidate when its anchor similarity fails to drop from the prior layer,
\begin{equation}
\mathcal{C} = \left\{\, l \;\middle|\; \bar{\boldsymbol{a}}^{(l)} - \bar{\boldsymbol{a}}^{(l-1)} \geq 0 \,\right\},
\end{equation}
where layers in \(\mathcal{C}\) introduce no distinct additional representation transformation. Candidates are ranked and filtered against a small protected set \(\mathcal{P}\) of the globally most fidelity‑critical layers, which are vetoed from pruning. The surviving top‑ranked candidates constitute the final pruning set,
\begin{equation}
\mathcal{S} = \left\{\, l \in \mathcal{C} \setminus \mathcal{P} \,\right\}, \qquad |\mathcal{S}| \leq K,
\end{equation}
bounded by a fixed pruning budget K. Selected layers are not pruned abruptly: early in training they undergo a temporary windowed freeze, later transitioning to a permanent freeze once the redundancy signal stabilizes, with a soft‑decay fallback when the pruning budget is saturated. This graduated schedule keeps LoRA‑SP conservative at early training stages while achieving persistent capacity reduction over time, enabling improved generation without severe fidelity degradation.

\subsection{Training Objective}

Following ~\cite{fang2026one}, one-step LoRA fine-tuning paradigm is adopted. As shown in Fig.~\ref{fig: main}, we replace the original Gaussian noise with LQ image as the input. The overall backbone generates fine-grained details in pure pixel-space by predicting the residual between LQ and GT image.

We fix pretrained DiT blocks, optimize LoRA with a reconstruction loss and an adversarial loss, yielding the total objective \(\mathcal{L} = \mathcal{L}_{\mathrm{rec}} + \mathcal{L}_{\mathrm{adv}}\). \(\mathcal{L}_{\mathrm{rec}}\) aggregates four terms: pixel-space regressive loss \(\mathcal{L}_1\), a DINOv3-ConvNeXt-based  ~\cite{simeoni2025dinov3} perceptual loss \(\mathcal{L}_{per}\), frequency-space loss~\cite{wang2026spectral} \(\mathcal{L}_{freq}\), and the InfoNCE contrastive loss \(\mathcal{L}_{\mathrm{InfoNCE}}\) (Sec.~\ref{sec:Distillation}).
\begin{equation}
\mathcal{L}_{\mathrm{rec}} = \lambda_{1}\mathcal{L}_{1} + \lambda_{\mathrm{2}}\mathcal{L}_{per} + \lambda_{3}\mathcal{L}_{freq} + \lambda_{4}\mathcal{L}_{InfoNCE}.
\label{eq:rec}
\end{equation}
For $\mathcal{L}_{\mathrm{adv}}$, we employ two complementary discriminators built on the same DINOv3-ConvNeXt features: a main discriminator $D_{\mathrm{main}}$ that performs multi-scale assessment of structural realism, guiding the model to produce images with higher visual authenticity, and our Cross-Scale Wavelet Discriminator $D_{\mathrm{wave}}$ (Sec.~\ref{sec:cswd}), which applies a wavelet transform to retain accurate high-frequency sub-bands, while penalizing texture/edge artifacts, especially boundary artifacts across patches:
\begin{equation}
\mathcal{L}_{\mathrm{adv}} = \lambda_{\mathrm{main}}\mathbb{E}_{\hat{x}}\big[-\log\sigma(D_{\mathrm{main}}(\hat{x}))\big]
+ \lambda_{\mathrm{wave}}(t)\,\mathbb{E}_{\hat{x}}\big[-\log\sigma(D_{\mathrm{wave}}(\hat{x}))\big],
\label{eq:adv}
\end{equation}
where $\lambda_{\mathrm{wave}}(t)$ is linearly warmed up to avoid the high-frequency term saturating at an early stage.

\subsection{Construction of Dual-Q-Bench}
\label{sec:construction in main}
Due to page limitations, please refer to Appendix~\ref{app:dual-q-bench} for construction details of Dual-Q-Bench.


\paragraph{User Study of Real-ISR Benchmarks.} We conduct a user study to evaluate the quality of Dual-Q-Bench against existing Real-ISR benchmarks, including RealSR~\cite{cai2019toward}, DRealSR~\cite{wei2020component}, GGT~\cite{kong2026ggt}, and ScreenSR~\cite{wu2026vosr}. We randomly sample $50$ GT-LQ image pairs from each dataset for evaluation. In each trial, $17$ participants are presented with five image pairs from different datasets and asked to select the best one based on GT quality (clarity, background de-focus, and realism), LQ authenticity, and GT–LQ alignment (color, brightness, and position). As shown in Tab.~\ref{tab:user_study_testsets}, Dual-Q-Bench achieves the highest user preference rate of $76\%$, substantially surpassing the second-best, GGT ($10.7\%$), confirming the best quality.


    

\begin{table*}[t]
  \vspace{-3mm}
  \setlength\tabcolsep{4pt}
  \setlength{\aboverulesep}{0.5pt}
  \setlength{\belowrulesep}{0.5pt}
  \caption{Quantitative results on real-world benchmarks, with the best/second-best results highlighted in \textbf{bold}/\underline{underlined}. "-s" denotes the number of diffusion steps.} 
  \label{tab:Quantitative}
  \vspace{-3mm}
  \resizebox{\textwidth}{!}{ 
  \centering
  \begin{tabular}{c|c|ccccc|ccccc|c}
    \toprule[2pt]
    Datasets & Metrics & \makecell{SUPIR\\-s50} & \makecell{FaithDiff\\-s20} & \makecell{Dit4SR\\-s40} & \makecell{VOSR\\-s25} & \makecell{LucidFlux\\-s15} & \makecell{TSDSR\\-s1} & \makecell{PiSA-SR\\-s1} & \makecell{TVT\\-s1} & \makecell{ODTSR\\-s1} & \makecell{StrSR\\-s1}  & \makecell{\textbf{Ours}\\-s1} \\
    \midrule \midrule
    \multirow{10}{*}{DrealSR} 
     & PSNR$\uparrow$    & 26.87   & 27.29  & 25.79  & 27.72  & 20.82  & 27.71  & 28.31  & \underline{28.25}  & 28.15  & 23.90  & \textbf{29.62} \\
     & SSIM$\uparrow$  & 0.6702  & 0.7104 & 0.6817 & 0.7376 & 0.5806 & 0.7543 & 0.7804 & \underline{0.7898} & 0.7740 & 0.6948  & \textbf{0.7990} \\
     & LPIPS$\downarrow$ & 0.4339  & 0.3497 & 0.3656 & 0.3664 & 0.4532 & 0.2997 & 0.2960 & 0.2900 & \textbf{0.2591} & 0.3345  & \underline{0.2898} \\
     & DISTS$\downarrow$ & 0.2793  & 0.2386 & 0.2437 & 0.2446 & 0.2771 & 0.2153 & 0.2169 & \underline{0.2206}  & \textbf{0.1926} & 0.2377 & 0.2302 \\
     & MANIQA$\uparrow$  & 0.5281  & \underline{0.6378} & 0.6232 & 0.5204 & 0.251 & 0.5883 & 0.6161 & 0.5784 & 0.6226 & 0.5598  & \textbf{0.6429} \\
     & MUSIQ$\uparrow$   & 53.91   & 65.79  & 64.51  & 57.70  & \underline{67.97}  & 66.83  & 66.11  & 65.60  & 62.83  & 60.98  & \textbf{68.18} \\
     & Q-Align$\uparrow$ & 3.2002  & 3.4284 & 3.3482 & 2.9609 & \underline{3.6537} & 3.6185 & 3.5836 & 3.6424 & 3.4652 & 3.3464 & \textbf{3.7473} \\
     & LIQE$\uparrow$    & 3.8725  & 3.9620 & 3.9157 & 2.7986 & 4.0107 & 3.9491 & 4.1496 & \underline{4.0861} & 3.5717 & 3.4685  & \textbf{4.1957} \\
     & Q-Insight$\uparrow$    & 3.1085  & 3.5806 & 3.5530 & 3.2112 & \textbf{3.6777} & 3.4693 & 3.5460 & 3.5133 & 3.4223 & 3.3051 & \underline{3.6419} \\
    & AFINE-ALL$\downarrow$ & 46.27  & 26.28 & 15.41 & 50.63  & \underline{14.45}  & 15.86  & 18.57  & 22.97  & 20.17  & 24.58  & \textbf{7.61} \\
    \midrule
    
    \multirow{10}{*}{\makecell{GGT-\\GeneralMix}} 
     & PSNR$\uparrow$    & \underline{26.03}  & 26.57  & 24.40  & 25.52  & 21.83  & 25.16  & 26.13  & 26.41  & 25.95  & 24.17 & \textbf{26.86} \\
     & SSIM$\uparrow$  & 0.6606  & 0.6984 & 0.6518 & 0.6512 & 0.5823 & 0.6813 & 0.7161 & \underline{0.7362} & 0.7126 & 0.6572 & \textbf{0.7435} \\
     & LPIPS$\downarrow$ & 0.3637  & 0.2800 & 0.3527 & 0.4418 & 0.4275 & 0.3114 & 0.2916 & \underline{0.2732} & 0.3040 & 0.3382 & \textbf{0.2684} \\
     & DISTS$\downarrow$ & 0.1945  & \underline{0.1614} & 0.1864 & 0.2750 & 0.2159 & 0.1701 & 0.1756 & 0.1631 & 0.1693 & 0.1846 & \textbf{0.1605} \\
     & MANIQA$\uparrow$  & 0.5677  & 0.6041 & 0.6068 & 0.4708 & 0.6087 & 0.5880 & 0.5934 & 0.5698 & \underline{0.6195} & 0.5597 & \textbf{0.6221} \\
     & MUSIQ$\uparrow$   & 59.36  & 63.13  & 68.49  & 49.79  & \underline{70.83}  & 69.27  & 65.60  & 67.79  & 66.37 & 62.18 & \textbf{71.06} \\
     & Q-Align$\uparrow$ & 3.7762  & 3.8415 & 3.9934 & 2.8977 & \underline{4.2334} & 3.8949 & 3.8818 & 3.9229 & 3.9480 & 3.9615 & \textbf{4.3009} \\
     & LIQE$\uparrow$    & 3.1497  & 3.3761 & 3.9904 & 1.9933 & 3.8951 & \underline{4.0377} & 3.7018 & 3.9262 & 3.6083 & 3.4264 & \textbf{4.2130} \\
     & Q-Insight$\uparrow$    & 3.4435  & 3.6085 & 3.8388 & 2.9985 & \textbf{3.9863} & 3.6901 & 3.6655 & 3.7361 & 3.7062 & 3.5839 & \underline{3.9289} \\
     & AFINE-ALL$\downarrow$ & 39.66  & 32.82 & 18.46 & 60.49 & 24.57 & 17.30 & 15.16  & 24.84  & \underline{12.40} & 26.32 & \textbf{2.6303} \\
     \midrule
     
     \multirow{10}{*}{\makecell{Dual-Q-\\Bench}} 
     & PSNR$\uparrow$    & \underline{22.71}  & 22.59 & 21.67 & 21.90 & 19.56 & 21.89 & 22.32 & 22.64 & 22.35 & 21.74 & \textbf{23.21} \\
     & SSIM$\uparrow$  & 0.6225  & 0.6196 & 0.6152 & 0.5701 & 0.5594 & 0.6290 & 0.6380 & 0.6492 & \underline{0.6507} & 0.6129 & \textbf{0.6662} \\
     & LPIPS$\downarrow$ & 0.3046  & 0.2736 & 0.3157 & 0.4095 & 0.3191 & 0.2648 & \underline{0.2722} & 0.2746 & 0.2862 & 0.3069 & \textbf{0.2639} \\
     & DISTS$\downarrow$ & 0.1784  & 0.1695 & 0.1793 & 0.2824 & 0.1829 & 0.1663 & 0.1675 & 0.1662 & \underline{0.1646} & 0.1788 & \textbf{0.1590} \\
     & MANIQA$\uparrow$  & 0.6257  & 0.6590 & 0.6328 & 0.4830 & \underline{0.6623} & 0.6307 & 0.6520 & 0.6250 & 0.6611 & 0.6061 & \textbf{0.6663} \\
     & MUSIQ$\uparrow$   & 67.80  & 70.80 & 70.87 & 59.12 & \underline{74.05} & 73.83 & 72.17 & 73.07 & 71.19 & 70.25 & \textbf{74.22} \\
     & Q-Align$\uparrow$ & 4.2362  & 4.2789 & 4.2239 & 2.8755 & \underline{4.5148} & 4.2297 & 4.3351 & 4.3144 & 4.3569 & 4.2257 & \textbf{4.6683} \\
     & LIQE$\uparrow$    & 4.1563  & 4.3841 & 4.3776 & 2.3265 & 4.4996 & \underline{4.6336} & 4.5401 & 4.6342 & 4.4962 & 4.3966 & \textbf{4.8425} \\
     & Q-Insight$\uparrow$    & 3.8389  & 3.9413 & 3.9807 & 3.2694 & \textbf{4.1875} & 3.9661 & 4.0004 & 4.0005 & 3.9700 & 4.0608 & \underline{4.1103} \\
     & AFINE-ALL$\downarrow$ & 47.01  & 43.44 & 41.68 & 69.75 & 42.64 & 39.00 & 35.10 & 40.22 & \underline{30.95} & 43.59 & \textbf{12.62} \\
    \bottomrule[2pt]
  \end{tabular}
  }
  \vspace{-3mm}
\end{table*}

\begin{table*}[t]
  \setlength\tabcolsep{6pt}
\setlength{\aboverulesep}{0.5pt}
  \setlength{\belowrulesep}{0.5pt}
  \caption{Comparison of computational costs on GGT-GeneralMix test set. }
  \label{tab:inference}
  \vspace{-3mm}
  \resizebox{\textwidth}{!}{ 
  \centering
  \begin{tabular}{c|ccccc|ccccc|c}
    \toprule[2pt]
     Methods & \makecell{SUPIR\\-s50} & \makecell{FaithDiff\\-s20} & \makecell{Dit4SR\\-s40} & \makecell{VOSR-1.4B\\-s25} & \makecell{LucidFlux\\-s15} & \makecell{TSDSR\\-s1} & \makecell{PiSA-SR\\-s1} & \makecell{TVT\\-s1} & \makecell{ODTSR\\-s1} & \makecell{StrSR\\-s1} & \makecell{\textbf{Ours}\\-s1} \\
    \midrule \midrule
     Inference Time (s) $\downarrow$ & 10.8  & 5.0 & 57.6 & 7.9 & 23.8 & 0.48 & 0.45 & 2.4 & 5.2 & 0.97  & \textbf{0.27} \\
     GPU Memory (G) $\downarrow$ & 52.46  & 11.10 & 22.06 & \underline{6.91} & 35.0 & 6.99 & \textbf{6.89} & 14.62 & 40.83 & 27.69  & 17.63 \\
    \bottomrule[2pt]
  \end{tabular}
  }
  \vspace{-7mm}
\end{table*}

\begin{figure*}[t]
	\begin{overpic}[width=0.99\linewidth]{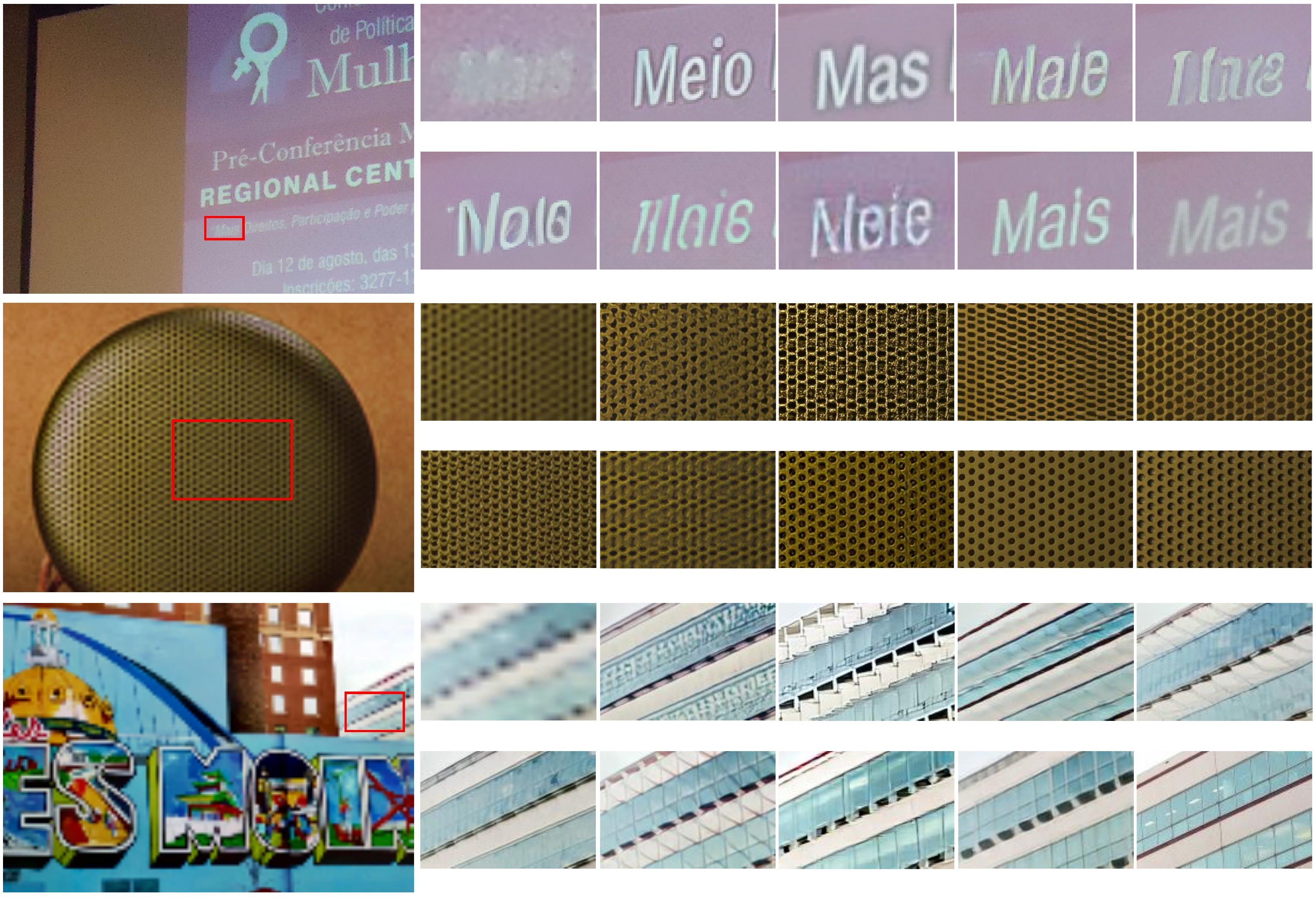}
    \put(37.4,57.3){\color{black}{\fontsize{8pt}{1pt}\selectfont LQ}}
    \put(49.4,57.3){\color{black}{\fontsize{8pt}{1pt}\selectfont Dit4SR}}
    \put(61.6,57.3){\color{black}{\fontsize{8pt}{1pt}\selectfont LucidFlux}}
    \put(76.1,57.3){\color{black}{\fontsize{8pt}{1pt}\selectfont TSDSR}}
    \put(88.7,57.3){\color{black}{\fontsize{8pt}{1pt}\selectfont PiSA-SR}}
    \put(36.4,46.1){\color{black}{\fontsize{8pt}{1pt}\selectfont TVT}}
    \put(48.4,46.1){\color{black}{\fontsize{8pt}{1pt}\selectfont ODTSR}}
    \put(63.4,46.1){\color{black}{\fontsize{8pt}{1pt}\selectfont StrSR}}
    \put(73.0,46.1){\color{black}{\fontsize{8pt}{1pt}\selectfont \textbf{HiFiPix (Ours)}}}
    \put(91.4,46.1){\color{black}{\fontsize{8pt}{1pt}\selectfont GT}}
    
	\put(37.4,34.6){\color{black}{\fontsize{8pt}{1pt}\selectfont LQ}}
    \put(49.4,34.6){\color{black}{\fontsize{8pt}{1pt}\selectfont Dit4SR}}
    \put(61.6,34.6){\color{black}{\fontsize{8pt}{1pt}\selectfont LucidFlux}}
    \put(76.1,34.6){\color{black}{\fontsize{8pt}{1pt}\selectfont TSDSR}}
    \put(88.7,34.6){\color{black}{\fontsize{8pt}{1pt}\selectfont PiSA-SR}}
    \put(36.4,23.3){\color{black}{\fontsize{8pt}{1pt}\selectfont TVT}}
    \put(48.4,23.3){\color{black}{\fontsize{8pt}{1pt}\selectfont ODTSR}}
    \put(63.4,23.3){\color{black}{\fontsize{8pt}{1pt}\selectfont StrSR}}
    \put(73.0,23.3){\color{black}{\fontsize{8pt}{1pt}\selectfont \textbf{HiFiPix (Ours)}}}
    \put(91.4,23.3){\color{black}{\fontsize{8pt}{1pt}\selectfont GT}}

    \put(37.4,11.8){\color{black}{\fontsize{8pt}{1pt}\selectfont LQ}}
    \put(49.4,11.8){\color{black}{\fontsize{8pt}{1pt}\selectfont Dit4SR}}
    \put(61.6,11.8){\color{black}{\fontsize{8pt}{1pt}\selectfont LucidFlux}}
    \put(76.1,11.8){\color{black}{\fontsize{8pt}{1pt}\selectfont TSDSR}}
    \put(88.7,11.8){\color{black}{\fontsize{8pt}{1pt}\selectfont PiSA-SR}}
    \put(36.4,0.3){\color{black}{\fontsize{8pt}{1pt}\selectfont TVT}}
    \put(48.4,0.3){\color{black}{\fontsize{8pt}{1pt}\selectfont ODTSR}}
    \put(63.4,0.3){\color{black}{\fontsize{8pt}{1pt}\selectfont StrSR}}
    \put(73.0,0.3){\color{black}{\fontsize{8pt}{1pt}\selectfont \textbf{HiFiPix (Ours)}}}
    \put(91.4,0.3){\color{black}{\fontsize{8pt}{1pt}\selectfont GT}}
    
	\end{overpic}
	\vspace{-1mm}
	\caption{Qualitative comparisons with existing Real-ISR methods on real-world test sets.}
	\label{fig: Qualitative}
	\vspace{-3mm}
\end{figure*}

\vspace{-2mm}
\section{Experiments}
\vspace{-2mm}
\subsection{Experimental Setup}
\vspace{-2mm}
\textbf{Datasets and Evaluation Protocols.}
We use LSDIR~\cite{li2023lsdir} and the first $10,000$ images from FFHQ~\cite{kornblith2019similarity} as the training sets. LQ images are synthesized using the degradation pipeline in Real-ESRGAN~\cite{wang2021real}. We evaluate competing methods on DRealSR~\cite{wei2020component}, GGT-GeneralMix~\cite{kong2026ggt}, RealSR~\cite{cai2019toward}  and our constructed \emph{Dual-Q-Bench}. To comprehensively evaluate all methods, we employ both full-reference (FR) and no-reference (NR) metrics. The PSNR, SSIM~\cite{wang2004image}, LPIPS~\cite{zhang2018unreasonable} and DISTS are used for fidelity evaluation, while MANIQA~\cite{yang2022maniqa}, MUSIQ~\cite{ke2021musiq}, Q-Insight~\cite{li2026q}, Q-Align~\cite{wu2023q}, and LIQE~\cite{zhang2023blind} are computed for visual generation assessment. In addition, we adopt AFINE-ALL~\cite{chen2025toward} to assess the consistency between full-reference and no-reference evaluation even the reference image is imperfect. 


\vspace{-1mm}
\textbf{Implementation Details.}
LoRA~\cite{hu2022lora} layers with rank $64$ are fine-tuned based on the open-source HiDream-O1-Image~\cite{hidreamolimage} model. $\lambda_1$, $\lambda_2$, $\lambda_3$, $\lambda_4$, $\lambda_{main}$, $\lambda_{wave}$ is set to $1$, $1$, $1e^{-1}$, $1$, $1$, $1e^{-1}$, respectively. The timestep $t$ is fixed to $0.5$ in training and inference stages. We train HiFiPix on $4$ NVIDIA H20 GPUs for $4,000$ iterations with an accumulated batch size of $32$. Following previous methods \cite{feilucidflux,wang2026spectral}, the training patch size is set to $1024^2$. The learning rate is set to $5e^{-5}$ with the AdamW \cite{loshchilov2019decoupled} optimizer. We discard image captions and remove the text encoder to accelerate training and inference.

\begin{table*}[t]
\centering
\begin{minipage}[t]{0.54\textwidth}
  \centering
  \raisebox{-8.5mm}{%
    \begin{overpic}[width=\linewidth]{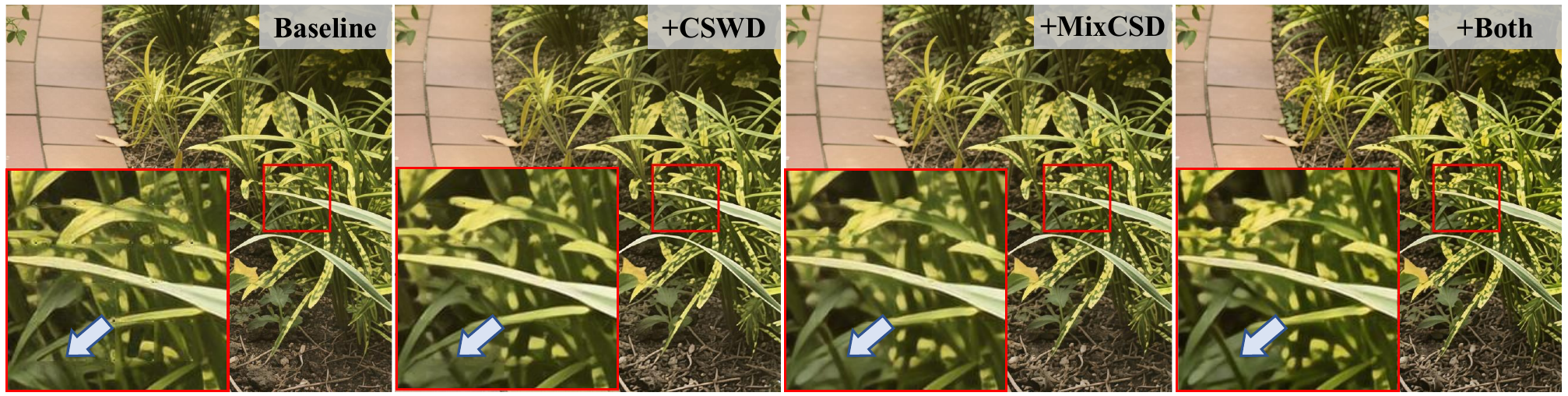}
    \end{overpic}
  }
  \vspace{-1mm}
  \captionof{figure}{Qualitative visualization of CSWD and MixCSD for patch‑boundary artifact mitigation.}
  \label{fig:patch_vis}
\end{minipage}
\hfill
\begin{minipage}[t]{0.44\textwidth}
  \centering
  \setlength\tabcolsep{2pt}
  \setlength{\aboverulesep}{0.5pt}
  \setlength{\belowrulesep}{0.5pt}
  \renewcommand{\arraystretch}{1.1}
  \resizebox{\linewidth}{!}{
  \begin{tabular}{cc|cccc}
      \toprule[2pt]
      CSWD & MixCSD & PSNR$\uparrow$ & LPIPS$\downarrow$ & MUSIQ$\uparrow$ & MANIQA$\uparrow$ \\
      \midrule \midrule
      \XSolidBrush & \XSolidBrush & 25.53 & 0.3132 & \textbf{70.67} & \textbf{0.6247} \\
      \Checkmark & \XSolidBrush   & 25.83 & 0.2781 & 68.62 & 0.6048  \\
      \XSolidBrush & \Checkmark   & 25.77 & 0.2839 & 68.88 & 0.6084   \\
      \Checkmark & \Checkmark     & \textbf{26.12} & \textbf{0.2759} & \underline{69.86} & \underline{0.6183}   \\
      \bottomrule[2pt]
  \end{tabular}}
  \vspace{-1mm}
  \caption{Ablation results of CSWD and MixCSD on quantitative metrics.}
  \label{tab:cswd_mixcsd_ablation}
\end{minipage}
\end{table*}
\vspace{-4mm}

\begin{table*}[t]
\setlength\tabcolsep{2pt}
  \setlength{\aboverulesep}{0.5pt}
  \setlength{\belowrulesep}{0.5pt}
\begin{minipage}{0.49\linewidth}
  \centering
  \caption{Ablation on the teacher network input composition: mixed LQ-HQ vs.\ HQ-only.}
  \vspace{-2mm}
  \scalebox{0.67}{
  \begin{tabular}{l|c|c|c|c}
    \toprule[1.5pt]
    \multicolumn{1}{l|}{\multirow{-1}{*}{Method}}
    & \multicolumn{1}{c|}{\multirow{-1}{*}{PSNR$\uparrow$}}
    & \multicolumn{1}{c|}{\multirow{-1}{*}{LPIPS$\downarrow$}}
    & \multicolumn{1}{c|}{\multirow{-1}{*}{MUSIQ$\uparrow$}}
    & \multicolumn{1}{c}{\multirow{-1}{*}{MANIQA$\uparrow$}}
    \\
    \midrule \midrule
    HQ-only                  & \textbf{26.15}  & \textbf{0.2726}  & \underline{68.72} & \underline{0.6079} \\
    LQ+HQ (\textbf{Ours})    & \underline{26.12} & \underline{0.2759} & \textbf{69.86} & \textbf{0.6183} \\
    \bottomrule[2pt]
  \end{tabular}}
  \label{tab:dis_input}
\end{minipage}
\hfill
\begin{minipage}{0.49\linewidth}
  \centering
  \caption{Ablation on loss formulation:  distillation loss ($\mathcal{L}_{\text{KD}}$) vs.\ contrastive learning ($\mathcal{L}_{\text{InfoNCE}}$).}
  \vspace{-2mm}
  \scalebox{0.67}{
  \begin{tabular}{l|c|c|c|c}
    \toprule[1.5pt]
    \multicolumn{1}{l|}{\multirow{-1}{*}{Method}}
    & \multicolumn{1}{c|}{\multirow{-1}{*}{PSNR$\uparrow$}}
    & \multicolumn{1}{c|}{\multirow{-1}{*}{LPIPS$\downarrow$}}
    & \multicolumn{1}{c|}{\multirow{-1}{*}{MUSIQ$\uparrow$}}
    & \multicolumn{1}{c}{\multirow{-1}{*}{MANIQA$\uparrow$}}
    \\
    \midrule \midrule
    $\mathcal{L}_{\text{KD}}$ & 26.01 & 0.2794 & 69.51 & 0.6155 \\
    $\mathcal{L}_{\text{InfoNCE}} (\textbf{Ours})$ & \textbf{26.12} & \textbf{0.2759} & \textbf{69.86} & \textbf{0.6183} \\
    \bottomrule[1.5pt]
  \end{tabular}}
  \label{tab:dis_loss}
\end{minipage}
\vspace{-2mm}
\end{table*}


\begin{table*}[t]
\setlength\tabcolsep{2pt}
  \setlength{\aboverulesep}{0.5pt}
  \setlength{\belowrulesep}{0.5pt}
\begin{minipage}{0.25\linewidth}
  \centering
  \caption{Results of user study, showing vote percentages for different Real-ISR benchmarks.}
  \vspace{-2mm}
  \scalebox{0.67}{
  \begin{tabular}{l|c}
    \toprule[1.5pt]
    \multicolumn{1}{l|}{\multirow{-1}{*}{Benchmarks}}
    & \multicolumn{1}{c}{\multirow{-1}{*}{Preferences$\uparrow$}}
    \\
    \midrule \midrule
    RealSR                  & 1.3\%  \\
    DrealSR                 & 1.5\%  \\
    GGT-GeneralMix          & 10.7\%  \\
    ScreenSR                & 10.5\%  \\
    Dual-Q-Bench (\textbf{Ours})  & 76.0\%  \\
    \bottomrule[1.5pt]
  \end{tabular}}
  \label{tab:user_study_testsets}
\end{minipage}
\hfill
\begin{minipage}{0.7\linewidth}
  \centering
  \caption{Effectiveness of redundancy detection and fidelity-aware veto in LoRA-SP. Further effectiveness of LoRA-SP in VAE-based models.}
  \vspace{-2mm}
  \scalebox{0.8}{
\begin{tabular}{c|c|c|c|c}
\toprule[1.5pt]
\multicolumn{1}{c|}{\multirow{-1}{*}{Method}}
& \multicolumn{1}{c|}{\multirow{-1}{*}{PSNR$\uparrow$}}
& \multicolumn{1}{c|}{\multirow{-1}{*}{LPIPS$\downarrow$}}
& \multicolumn{1}{c|}{\multirow{-1}{*}{MUSIQ$\uparrow$}}
& \multicolumn{1}{c}{\multirow{-1}{*}{MANIQA$\uparrow$}}
\\
\midrule \midrule
Hidream Baseline (Pixel-based)                           & \textbf{26.12} & \textbf{0.2759} & 69.86 & 0.6183 \\
+Random LoRA Pruning                    & 25.27 & 0.2910 & 69.52 & 0.6124 \\
+Redundancy detection (RD)          & 25.81 & 0.2846 & \textbf{71.17} & \textbf{0.6303} \\
+RD \& Fidelity-aware veto (\textbf{Ours})          & \underline{26.06} & \underline{0.2775} & \underline{70.92} & \underline{0.6274} \\
\hline
Z-Image (VAE-based)    & \textbf{23.81}       & \textbf{0.6327}     & \underline{72.21}        & \textbf{0.6293}    \\ 
Z-Image+LoRA-SP (VAE-based)    & \underline{23.05}       & \underline{0.6205}     & \textbf{72.72}        & \underline{0.6267}    \\ 
\bottomrule[1.5pt]
\end{tabular}}
\label{tab:lora_sp}
\end{minipage}
\vspace{-6mm}
\end{table*}

\vspace{-4mm}
\subsection{Experiment Results}
\label{sec:exp result in main}
\vspace{-4mm}
We compare HiFiPix with existing state-of-the-art Real-ISR methods, including multi-steps methods (\ie, SUPIR~\cite{yu2024scaling}, FaithDiff~\cite{chen2025faithdiff}, Dit4SR~\cite{duan2025dit4sr}, VOSR~\cite{wu2026vosr} in version of 1.4B Params,  LucidFlux~\cite{feilucidflux}), and one-step methods (\ie, TSDSR~\cite{dong2025tsd}, PiSA-SR~\cite{sun2025pixel}, TVT~\cite{yi2025fine}, ODTSR~\cite{fang2026one}, and StrSR~\cite{wang2026spectral} based on Z-Image-Turbo~\cite{cai2025z}).

\textbf{Quantitative Results.}
As shown in Tab.~\ref{tab:Quantitative}, HiFiPix consistently outperforms existing methods in restoration fidelity, with notable gains on high-quality GT benchmarks (GGT and Dual-Q-Bench). It also excels on most NR metrics, confirming enhanced fidelity without sacrificing generation ability. 

\textbf{Qualitative Results.}
As shown in Fig.~\ref{fig: Qualitative}, HiFiPix produces both photo-realistic results and faithfully restored details, e.g., fine text in the first set and faithful textures in the second and third sets. By contrast, existing VAE-based methods generate entangled artifacts and over-smoothed details. 

\textbf{Computational Cost.} 
As shown in Tab.~\ref{tab:inference}, we compare the computational cost of different methods on GGT-GeneralMix at \(1024^2\) resolution on a single NVIDIA H20 GPU.  Benefiting from the removal of the text encoder and the efficient vanilla patch-wise tokenization in HiDream-O1-Image that greatly reduces complexity, HiFiPix is much more efficient than multi-step diffusion methods and runs 40\% faster than PiSA-SR, the second-best single-step method. Moreover, its GPU memory footprint is considerably smaller than ODTSR and StrSR, which also leverage large‑scale pre‑trained models. Due to page limitations, comparisons of parameters can be found in Tab.~\ref{tab: parameter} of Appendix A.3.

\subsection{Ablation Study}
\label{sec:abl study in main}
\vspace{-1mm}


\textbf{CSWD \& MixCSD.} As shown in Fig.~\ref{fig:patch_vis}, both CSWD and MixCSD effectively suppress periodic boundary artifacts, and their combination yields the best visual results. From Tab.~\ref{tab:cswd_mixcsd_ablation}, both terms also substantially improve reconstruction fidelity (PSNR/LPIPS), yet using either alone degrades generation quality (MUSIQ/MANIQA). In contrast, combining them achieves the best fidelity without excessive loss of generation capability. CSWD and MixCSD penalize artifacts from image and intermediate feature domains, respectively, making them complementary.

\textbf{GT‑LQ Mixing Strategy.} We conduct comparison experiments by replacing the GT‑LQ mixture branch with GT‑only inputs. Tab.~\ref{tab:dis_input} validates the benefits of mixed LQ‑HQ teacher inputs. An HQ‑only teacher creates a huge domain gap relative to LQ student inputs, degrading performance.

\textbf{Contrastive Loss.} We substitute contrastive loss with knowledge distillation loss~\cite{dong2025tsd}. As shown in Tab.~\ref{tab:dis_loss}, contrastive learning is superior to distillation by leveraging negative samples to avoid representation collapse and maintain feature diversity, thereby improving performance.

\textbf{LoRA‑SP.} LoRA‑SP identifies redundant LoRA layers for pruning and incorporates a fidelity‑aware veto mechanism, which retains critical layers by postponing their pruning to later candidates. As shown in Tab.~\ref{tab:lora_sp}, redundancy detection alone boosts generation quality yet causes notable fidelity drops. Equipped with the fidelity‑aware veto, LoRA‑SP yields an improved generation‑fidelity trade‑off and consistently surpasses uniform and random LoRA pruning. These results validate our design and suggest an interpretable path to enhance pixel‑space model generation via selective LoRA pruning. Moreover, Tab.~\ref{tab:lora_sp} indicates that this strategy yields marginal gains for latent‑space models (\eg, Z-Image), probably because VAE‑driven semantic compression produces little layer redundancy. Hence, LoRA‑SP is tailored for boosting the generation performance of pixel‑space models.

Due to page limitations, more ablation studies can be found in Appendix.~\ref{sec:abl study in supp}.


\subsection{Limitations}


HiFiPix suffers from two main limitations. First, while patch‑boundary artifacts are mostly mitigated, several extremely blurred inputs still contain faint residual seams within texture‑rich areas. Adding a generative decoder (\eg, PiD~\cite{lu2026pid}) could suppress these artifacts further, yet we abandon it for the additional consumption. Second, as our pipeline does not rely on text prompts, we exclude text‑conditioned metrics such as CLIPIQA. These metrics focus mainly on text‑image alignment, thereby our prompt‑free setting yields inherently low scores that offer limited evaluation results.

\section{Conclusion}
\vspace{-1mm}
We present HiFiPix, a pixel‑space generative framework which enables faithful restoration by removing VAE compression. To mitigate patch‑boundary artifacts and generative degradation, we propose CSWD, MixCSD for enhanced high‑frequency reconstruction and cross‑patch consistency, and LoRA‑SP to generate fine-grained details with minimal fidelity sacrifice. We further construct Dual‑Q‑Bench, a high‑quality benchmark for reliable evaluation. Experiments show HiFiPix outperforms VAE‑based methods, demonstrating the promise of pixel‑space modeling for Real‑ISR.

\subsection*{AI use statement}
In this work, we used generative AI tools to improve the readability, grammar, and language of the manuscript, and to assist with the creation and modification of software code. Generative AI tools were not used to develop the methodology, formulate mathematical claims, design experiments, or analyze or interpret experimental results. All AI-assisted text and code were reviewed, revised, and verified by the authors. We take full responsibility for the final content of this work, including all text, code, claims, and conclusions.

\bibliography{iclr2027_conference}
\bibliographystyle{iclr2027_conference}

\appendix
\section{Appendix}
In this Appendix file, we provide the following materials:
\begin{itemize}
    \item Construction of Dual-Q-Bench. Please refer to Sec.~\ref{sec:construction in main} in the main paper.
    \item More quantitative and qualitative comparisons. Please refer to Sec.~\ref{sec:exp result in main} in the main paper.
    \item Model parameters of different Real-ISR methods. Please refer to Sec.~\ref{sec:exp result in main} in the main paper.
    \item More ablation studies of our method. Please refer to Sec.~\ref{sec:abl study in main} in the main paper.
\end{itemize}

\subsection{Construction of Dual-Q-Bench}
\label{app:dual-q-bench}
\begin{figure*}[t]
\begin{overpic}[width=0.99\linewidth]{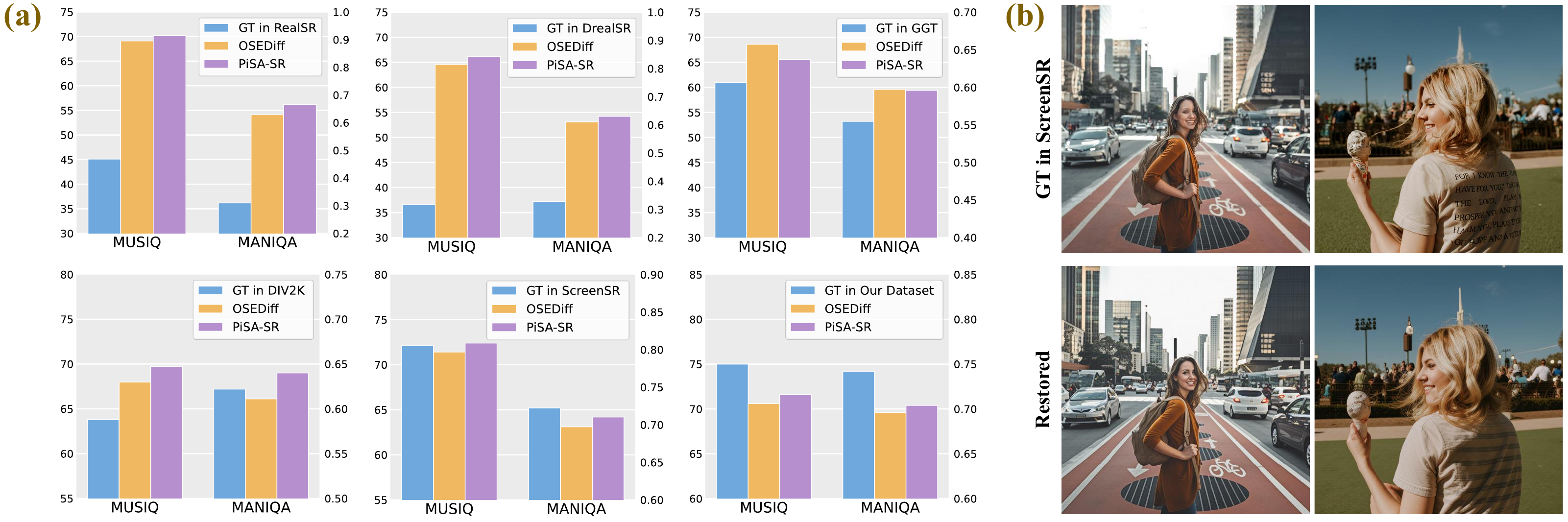}
\end{overpic}
\vspace{-0mm}
\caption{\textbf{(a) Insufficient GT Quality.} Existing Real-ISR benchmarks often provide GT with limited sharpness. Consequently, restored results may contain substantially richer or sharper details than GT, achieving higher MUSIQ and MANIQA scores than GT. This mismatch makes such GTs unreliable for jointly evaluating restoration fidelity and visual generation effects. \textbf{(b) Background De-focus in GT.} Existing benchmarks, such as ScreenSR, contain GT images with de-focused backgrounds. Real-ISR models often recover details in these regions that are absent from the GT, increasing the discrepancy with the reference and consequently challenging fidelity evaluation.}
\label{fig: benchmark}
\vspace{-0mm}
\end{figure*}

Existing paired Real-ISR benchmarks often suffer from low-quality or unrealistic GT, which can compromise the reliability of fidelity and generation evaluation. As shown in Fig.~\ref{fig: benchmark}, existing benchmarks (\eg, RealSR~\cite{cai2019toward} and DrealSR~\cite{wei2020component}) contain GT with blur details or background de-focus. Such imperfections introduce an undesirable evaluation bias: outputs with richer and more realistic details may deviate from deficient GT and receive lower fidelity scores, while outputs that closely mimic the imperfect GT may be favored instead, finally leading to unreasonable measurements, which is also indicated in ~\cite{chen2025toward}. Besides, background de-focus in DIV2K~\cite{agustsson2017ntire} and ScreenSR~\cite{wu2026vosr} complicates fidelity assessment, as recovering sharp structures in the background can be incorrectly penalized as deviation from the GT. Furthermore, large-model-generated GT (\eg, GGT-GeneralMix~\cite{kong2026ggt}) do not necessarily guarantee faithful details due to potential hallucinations. To address these limitations, we construct \emph{Dual-Q-Bench}, a high-quality real-world paired benchmark with realistic and sharp GT images at $1024^2$ resolution, enabling more reliable evaluation of both fidelity and generation quality.

\textbf{GT Construction:} We crawl high-resolution photographs from Unsplash and filter them using a combination of NR metrics (\eg, MANIQA~\cite{yang2022maniqa}, MUSIQ~\cite{ke2021musiq}, Q-Insight~\cite{li2026q}), retaining only images whose aggregated score exceeds GT images in LSDIR~\cite{li2023lsdir} by an average of 30\% to ensure superior intrinsic quality. Laplacian-gradient response is then used to crop texture-rich regions, followed by manual inspection to discard images with background de-focus or painterly appearance, yielding $100$ GT images with sharp, rich details at $1024^2$ resolution. 

\textbf{LQ Acquisition:} We obtain real-world LQ images following the same screen photographing strategy in ~\cite{wu2026vosr}. Specifically, the original GT images are shown on an ultra-high-definition screen in the Imaging Laboratory. We implement a timed routine to capture LQ from GT images displayed on‑screen by using a mobile phone for real‑world degradation simulation. The specific mobile phone model used for image capture will be disclosed upon acceptance of our paper.

\textbf{GT-LQ Alignment Strategy:} Each LQ is registered to its GT via SIFT keypoint matching with RANSAC homography estimation, falling back to multi-scale template matching when correspondences are insufficient, and further refined by ECC for sub-pixel accuracy. The warped LQ then undergoes locally-adaptive brightness and color correction in LAB space to compensate for capture-induced illumination shifts, after which both images are center-cropped to $1024^2$ to form the final aligned pair. Tab.~\ref{tab:datasets} shows the differences between Dual-Q-Bench and existing benchmarks.

\begin{table*}[t]
    \setlength\tabcolsep{8.5pt}
    \setlength{\aboverulesep}{0.4pt}
    \setlength{\belowrulesep}{0.4pt}
    \centering
    \caption{\small Comparisons between existing paired Real-ISR benchmarks and our Dual-Q-Bench.}
    \vspace{-2mm}
    \label{tab:datasets}
    \scalebox{0.75}{\begin{tabular}{c|c|c|cccc|c}
        \toprule[2pt]
        \multirow{2}{*}{Benchmark} & \# of & \multirow{2}{*}{Resolution} & MANIQA$\uparrow$ & MUSIQ$\uparrow$ & Real & Real  & Construction \\
        & Pairs & & of GT & of GT & LQ & GT  & Manner \\
        \midrule[1pt] \midrule[1pt]
        DIV2K & 100 & $512^2$ & 0.6274 & 63.7 & \XSolidBrush & \checkmark &  Synthesized LQ \\
        RealSR & 100 & $512^2$ & 0.6017 & 57.5 & \checkmark & \checkmark  & DSLR acquisition \\
        DrealSR & 100 & $512^2$ & 0.5589 & 50.1 & \checkmark & \checkmark  & DSLR acquisition \\
        ScreenSR & 130 & $512^2$ & 0.7187 & 72.2 & \checkmark & \checkmark  & Screen photography \\
        GGT-GeneralMix & 250 & $1024^2$ & 0.6488 & 68.3 & \checkmark & \XSolidBrush  & Generated GT \\
        \midrule[1pt]
        \textbf{Dual-Q-Bench (Ours)} & 100 & $1024^2$ & \textbf{0.7316} & \textbf{74.9} & $\checkmark$ & $\checkmark$  & Screen photography \\
        \bottomrule[2pt]
    \end{tabular}}
    \vspace{-1mm}
\end{table*}

\begin{table*}[t]
  \setlength\tabcolsep{4pt}
    \setlength{\aboverulesep}{0.4pt}
  \setlength{\belowrulesep}{0.4pt}
  \caption{Quantitative comparisons with existing Real-ISR methods on real-world benchmarks, with the best/second-best results in \textbf{bold}/\underline{underlined}. "-s" denotes the number of diffusion steps.}
  \label{tab:realsr}
  \resizebox{\textwidth}{!}{ 
  \centering
  \begin{tabular}{c|c|ccccc|ccccc|c}
    \toprule[2pt]
    Datasets & Metrics & \makecell{SUPIR\\-s50} & \makecell{FaithDiff\\-s20} & \makecell{Dit4SR\\-s40} & \makecell{VOSR\\-s25} & \makecell{LucidFlux\\-s15} & \makecell{TSDSR\\-s1} & \makecell{PiSA-SR\\-s1} & \makecell{TVT\\-s1} & \makecell{ODTSR\\-s1} & \makecell{StrSR\\-s1}  & \makecell{\textbf{Ours}\\-s1} \\
    \midrule \midrule
    \multirow{10}{*}{RealSR} 
     & PSNR$\uparrow$    & 25.09   & 25.28  & 23.39  & 25.29  & 18.58  & 24.80  & 25.50  & \underline{25.81}  & 25.07  & 21.66 & \textbf{26.70} \\
     & SSIM$\uparrow$    & 0.6728  & 0.7073 & 0.6661 & 0.7150 & 0.5544 & 0.7159 & 0.7418 & \textbf{0.7597} & 0.7361 & 0.6531 & \underline{0.7553} \\
     & LPIPS$\downarrow$ & 0.3731  & 0.2872 & 0.3193 & 0.2961 & 0.3856 & 0.2997 & 0.2672 & \underline{0.2593} & \textbf{0.2395} & 0.3126 & 0.2622 \\
     & DISTS$\downarrow$ & 0.2515  & 0.2110 & 0.2234 & 0.2226 & 0.2658 & 0.2153 & 0.2044 & \underline{0.2061} & \textbf{0.1892} & 0.2358 & 0.2191 \\
     & MANIQA$\uparrow$  & 0.5632  & \underline{0.6737} & 0.6569 & 0.6510 & 0.6645 & 0.6342 & 0.6551   & 0.6225 & 0.6620 & 0.5967 & \textbf{0.6739} \\
     & MUSIQ$\uparrow$   & 57.83   & 68.91  & 67.72  & 70.47  & \textbf{71.40}  & 71.23  & 70.15   & 69.86  & 68.29  & 66.15 & \underline{70.78} \\
     & Q-Align$\uparrow$ & 3.3511  & 3.5679 & 3.3780 & 3.2089 & \underline{3.8472} & 3.7364 & 3.6345  & 3.7679 & 3.6065 & 3.6615 & \textbf{3.9069} \\
     & LIQE$\uparrow$    & 2.9247  & 3.9748 & 3.9351 & 3.3794 & \textbf{4.2172} & 4.0649 & 4.0982  & \underline{4.1567} & 3.9295 & 3.6873 & \underline{4.2034} \\
    & AFINE-ALL$\downarrow$ & 46.55  & 20.86 & 15.38 & 30.68 & \underline{10.84}  & 17.45  & 16.87    & 23.79  & 15.07  & 28.37 & \textbf{6.32} \\

    \midrule
    \multirow{5}{*}{RealLQ250} 
     & MANIQA$\uparrow$  & 0.6547  & 0.6877  & 0.6832 & 0.6811 & 0.6832  & 0.6726  & 0.6912   & 0.6317  & \textbf{0.7103} & 0.6348 & \underline{0.6997} \\
     & MUSIQ$\uparrow$   & 69.55  & 71.61 & 72.00  & 71.52  & 73.80   & \textbf{74.16}   & 73.13   & 69.37   & 73.16  &71.35 & \underline{73.82} \\
     & Q-Align$\uparrow$ & \underline{4.3184}  & 4.0015 & 3.9845 & 3.8657 & 4.2847  & 4.1936  & 4.1849   & 4.0287  & 4.1307 &4.0631 & \textbf{4.4402} \\
     & LIQE$\uparrow$    & 4.3553  & 4.4080  & 4.4912 & 4.2704 & 4.6271  & \underline{4.6227}  & 4.6213   & 4.2508  & 4.5284 &4.2024 & \textbf{4.7264} \\
    
    \bottomrule[2pt]
  \end{tabular}
  }
\end{table*}

\begin{table*}[ht]
  \setlength\tabcolsep{3pt}
  \setlength{\aboverulesep}{0.4pt}
  \setlength{\belowrulesep}{0.4pt}
  \caption{Comparison of model parameters, where the generative model includes the text encoder, the VAE encoder/decoder, and the diffusion backbones; a language model is utilized for generating image captions. Since HiFiPix does not require image captions or text encoders, the final model parameters are fewer than the original $9B$ HiDream-O1-Image (with text encoder). }
  \vspace{-1mm}
  \resizebox{\textwidth}{!}{ 
  \centering
  \begin{tabular}{c|ccccc|ccccc|c}
    \toprule[2pt]
     Methods & \makecell{SUPIR\\-s50} & \makecell{FaithDiff\\-s20} & \makecell{Dit4SR\\-s40} & \makecell{VOSR\\-s25} & \makecell{LucidFlux\\-s15} & \makecell{TSDSR\\-s1} & \makecell{PiSA-SR\\-s1} & \makecell{TVT\\-s1} & \makecell{ODTSR\\-s1} & \makecell{StrSR\\-s1} & \makecell{\textbf{Ours}\\-s1} \\
    \midrule \midrule
     Generative Model params (B)$\downarrow$ & 4.80 & 3.51 & 8.38 & 1.74 & 14.03 & 7.79 & 1.29 & 1.26 & 21.86 & 6.24 & 7.77 \\
     Language Model Params (B)$\downarrow$ & 13.35 & 13.35 & 13.35 & 0 & 0 & 0 & 0 & 0.47 & 8.29 & 4.44 & 0 \\
     Trainable Parts Params (M)$\downarrow$ & 1243.4  & 2612.2 & 350.9 & 1439 & 1630.8 & 95.2  & 8.1 & 30.07 & 1231.2   & 641.6 & 207.1 \\
    \bottomrule[2pt]
  \end{tabular}
  }
  \label{tab: parameter}
  \vspace{-0mm}
\end{table*}

\begin{table*}[!h]
\setlength\tabcolsep{3pt}
  \setlength{\aboverulesep}{0.4pt}
  \setlength{\belowrulesep}{0.4pt}
\begin{minipage}{0.48\linewidth}
  \centering 
  \caption{Ablation on the number of selected layers in our LoRA-SP.}
  \vspace{-1mm}
    \scalebox{0.72}{
    \begin{tabular}{l|c|c|c|c}
\toprule[1.5pt]
    \multicolumn{1}{l|}{\multirow{-1}{*}{Layers}} 
    & \multicolumn{1}{c|}{\multirow{-1}{*}{PSNR$\uparrow$}}
    & \multicolumn{1}{c|}{\multirow{-1}{*}{LPIPS$\downarrow$}}
    & \multicolumn{1}{c|}{\multirow{-1}{*}{MUSIQ$\uparrow$}}
    & \multicolumn{1}{c}{\multirow{-1}{*}{MANIQA$\uparrow$}}

    \\ 
    \midrule \midrule
    1 layer    & 26,11      & 0.2764     & 70.15  & 0.6207   \\ 
    2 layers    & 26.09      & 0.2768     & 70.51  & 0.6236    \\ 
    4 layers (Ours) & 26.06 & 0.2775  & 70.92  & 0.6274    \\ 
    6 layers    & 25.92      & 0.2798     & 71.07  & 0.6282    \\ 
    8 layers    & 25.63      & 0.2856     & 71.22  & 0.6298    \\ 

    \bottomrule[1.5pt]
\end{tabular}}
\vspace{2.5mm}
\label{tab: num_lora}
\end{minipage}
~
\begin{minipage}{0.48\linewidth}
  \centering 
  \caption{Ablation on the number of distillation layers in MixCSD.}
  \vspace{-1mm}
    \scalebox{0.72}{
    \begin{tabular}{l|c|c|c|c}
\toprule[1.5pt]
    \multicolumn{1}{l|}{\multirow{-1}{*}{Layers}} 
    & \multicolumn{1}{c|}{\multirow{-1}{*}{PSNR$\uparrow$}}
    & \multicolumn{1}{c|}{\multirow{-1}{*}{LPIPS$\downarrow$}}
    & \multicolumn{1}{c|}{\multirow{-1}{*}{MUSIQ$\uparrow$}}
    & \multicolumn{1}{c}{\multirow{-1}{*}{MANIQA$\uparrow$}}

    \\ 
    \midrule \midrule
    3 layers    & 25.90      & 0.2804     &  69.37       & 0.6138    \\ 
    4 layers    & 26.02      & 0.2766     &  69.69  & 0.6168    \\ 
    5 layers (Ours) & 26.12 & \textbf{0.2759}  & \textbf{69.86} & \textbf{0.6183}    \\ 
    6 layers    & 26.16      & 0.2761     & 69.61 & 0.6152    \\ 
    7 layers    &  \textbf{26.21}     & 0.2767     & 69.15  & 0.6120    \\ 
    \toprule[1.5pt]
\end{tabular}}
    \vspace{2.5mm}
    \label{tab: num_dis}
\end{minipage}
\vspace{-5mm}
\end{table*}



\subsection{Quantitative and Qualitative Comparisons}
\paragraph{Quantitative Comparisons.}
As shown in Tab.~\ref{tab:realsr}, on existing test sets, HiFiPix achieves competitive pixel‑level restoration fidelity and obtains the best MANIQA~\cite{yang2022maniqa}, Q‑Align~\cite{wu2023q}, and AFINE‑ALL~\cite{chen2025toward} scores. Its relatively lower LPIPS~\cite{zhang2018unreasonable} and DISTS~\cite{ding2020image} scores stem partly from limited perceptual quality within GT images, which lack fine details that HiFiPix faithfully reconstructs, rendering these FR metrics unreliable. Overall, HiFiPix improves fidelity without degrading generative performance.

\begin{figure*}[t]
	\begin{overpic}[width=0.99\linewidth]{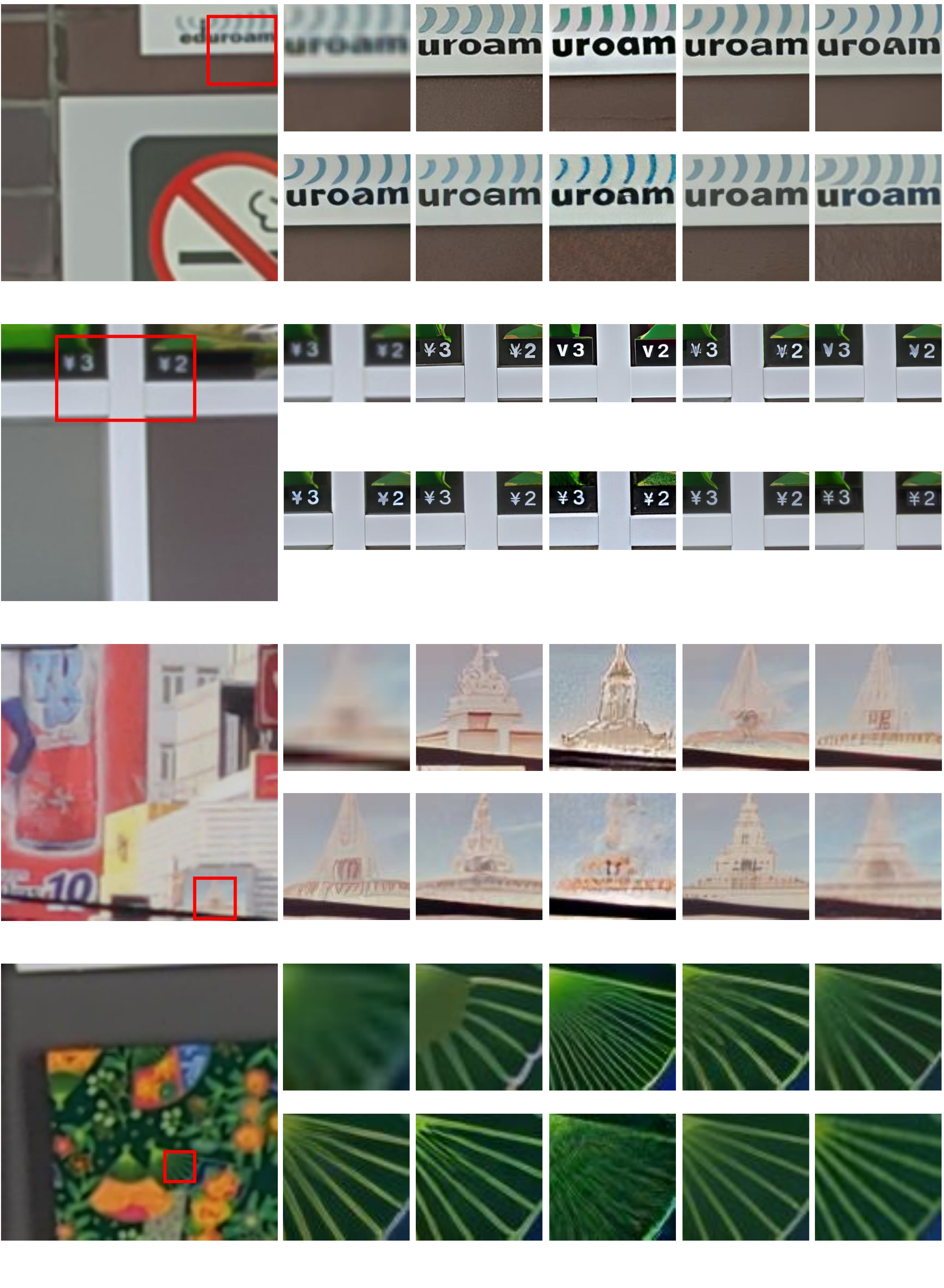}
    \put(26.0,88.6){\color{black}{\fontsize{8pt}{1pt}\selectfont LQ}}
    \put(35.0,88.6){\color{black}{\fontsize{8pt}{1pt}\selectfont Dit4SR}}
    \put(44.6,88.6){\color{black}{\fontsize{8pt}{1pt}\selectfont LucidFlux}}
    \put(55.1,88.6){\color{black}{\fontsize{8pt}{1pt}\selectfont TSDSR}}
    \put(65.4,88.6){\color{black}{\fontsize{8pt}{1pt}\selectfont PiSA-SR}}
    \put(25.5,77.0){\color{black}{\fontsize{8pt}{1pt}\selectfont TVT}}
    \put(35.0,77.0){\color{black}{\fontsize{8pt}{1pt}\selectfont ODTSR}}
    \put(45.4,77.0){\color{black}{\fontsize{8pt}{1pt}\selectfont StrSR}}
    \put(53.0,77.0){\color{black}{\fontsize{8pt}{1pt}\selectfont \textbf{HiFiPix (Ours)}}}
    \put(66.8,77.0){\color{black}{\fontsize{8pt}{1pt}\selectfont GT}}
    \put(6.8,77.0){\color{black}{\fontsize{8pt}{1pt}\selectfont From RealSR}}
    
    \put(26.0,67.1){\color{black}{\fontsize{8pt}{1pt}\selectfont LQ}}
    \put(35.0,67.1){\color{black}{\fontsize{8pt}{1pt}\selectfont Dit4SR}}
    \put(44.6,67.1){\color{black}{\fontsize{8pt}{1pt}\selectfont LucidFlux}}
    \put(55.1,67.1){\color{black}{\fontsize{8pt}{1pt}\selectfont TSDSR}}
    \put(65.4,67.1){\color{black}{\fontsize{8pt}{1pt}\selectfont PiSA-SR}}
    \put(25.5,55.7){\color{black}{\fontsize{8pt}{1pt}\selectfont TVT}}
    \put(35.0,55.7){\color{black}{\fontsize{8pt}{1pt}\selectfont ODTSR}}
    \put(45.4,55.7){\color{black}{\fontsize{8pt}{1pt}\selectfont StrSR}}
    \put(53.0,55.7){\color{black}{\fontsize{8pt}{1pt}\selectfont \textbf{HiFiPix (Ours)}}}
    \put(66.8,55.7){\color{black}{\fontsize{8pt}{1pt}\selectfont GT}}
    \put(6.8,52.0){\color{black}{\fontsize{8pt}{1pt}\selectfont From RealSR}}
    
	\put(26.0,38.9){\color{black}{\fontsize{8pt}{1pt}\selectfont LQ}}
    \put(35.0,38.9){\color{black}{\fontsize{8pt}{1pt}\selectfont Dit4SR}}
    \put(44.6,38.9){\color{black}{\fontsize{8pt}{1pt}\selectfont LucidFlux}}
    \put(55.1,38.9){\color{black}{\fontsize{8pt}{1pt}\selectfont TSDSR}}
    \put(65.4,38.9){\color{black}{\fontsize{8pt}{1pt}\selectfont PiSA-SR}}
    \put(25.5,27.2){\color{black}{\fontsize{8pt}{1pt}\selectfont TVT}}
    \put(35.0,27.2){\color{black}{\fontsize{8pt}{1pt}\selectfont ODTSR}}
    \put(45.4,27.2){\color{black}{\fontsize{8pt}{1pt}\selectfont StrSR}}
    \put(53.0,27.2){\color{black}{\fontsize{8pt}{1pt}\selectfont \textbf{HiFiPix (Ours)}}}
    \put(66.8,27.2){\color{black}{\fontsize{8pt}{1pt}\selectfont GT}}
    \put(6.8,27.2){\color{black}{\fontsize{8pt}{1pt}\selectfont From DrealSR}}

    \put(26.0,14.0){\color{black}{\fontsize{8pt}{1pt}\selectfont LQ}}
    \put(35.0,14.0){\color{black}{\fontsize{8pt}{1pt}\selectfont Dit4SR}}
    \put(44.6,14.0){\color{black}{\fontsize{8pt}{1pt}\selectfont LucidFlux}}
    \put(55.1,14.0){\color{black}{\fontsize{8pt}{1pt}\selectfont TSDSR}}
    \put(65.4,14.0){\color{black}{\fontsize{8pt}{1pt}\selectfont PiSA-SR}}
    \put(25.5,2.5){\color{black}{\fontsize{8pt}{1pt}\selectfont TVT}}
    \put(35.0,2.5){\color{black}{\fontsize{8pt}{1pt}\selectfont ODTSR}}
    \put(45.4,2.5){\color{black}{\fontsize{8pt}{1pt}\selectfont StrSR}}
    \put(53.0,2.5){\color{black}{\fontsize{8pt}{1pt}\selectfont \textbf{HiFiPix (Ours)}}}
    \put(66.8,2.5){\color{black}{\fontsize{8pt}{1pt}\selectfont GT}}
    \put(6.8,2.5){\color{black}{\fontsize{8pt}{1pt}\selectfont From DrealSR}}
    
	\end{overpic}
	\vspace{-1mm}
	\caption{Qualitative comparisons with existing Real-ISR methods on paired LQ-GT real-world test sets: RealSR and DrealSR.}
	\label{fig: Qualitative_supp1}
	\vspace{-3mm}
\end{figure*}

\begin{figure*}[t]
	\begin{overpic}[width=0.99\linewidth]{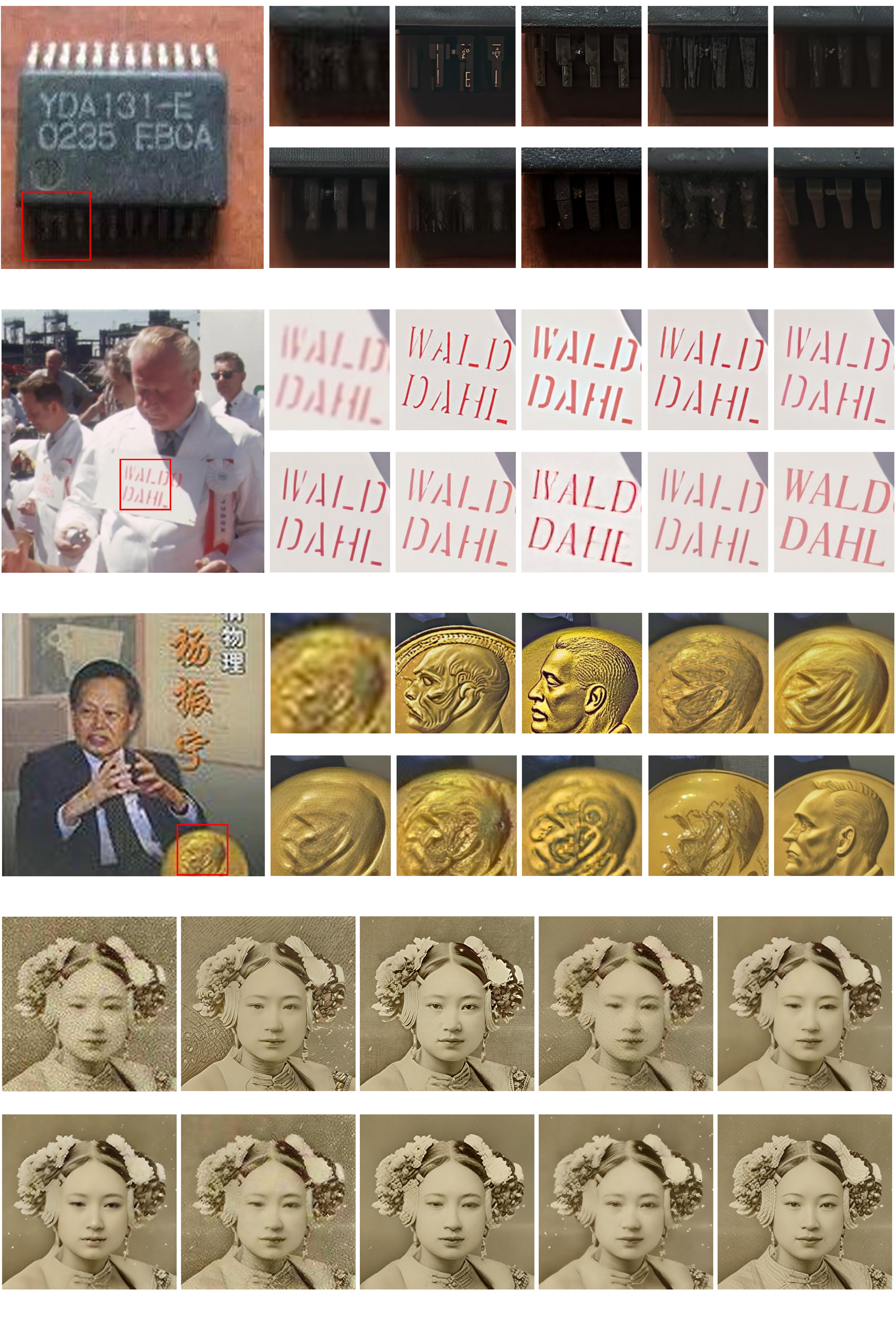}
    \put(24.0,89.3){\color{black}{\fontsize{8pt}{1pt}\selectfont LQ}}
    \put(32.6,89.3){\color{black}{\fontsize{8pt}{1pt}\selectfont Dit4SR}}
    \put(41.9,89.3){\color{black}{\fontsize{8pt}{1pt}\selectfont LucidFlux}}
    \put(51.8,89.3){\color{black}{\fontsize{8pt}{1pt}\selectfont TSDSR}}
    \put(60.8,89.3){\color{black}{\fontsize{8pt}{1pt}\selectfont PiSA-SR}}
    \put(23.6,78.5){\color{black}{\fontsize{8pt}{1pt}\selectfont TVT}}
    \put(32.4,78.5){\color{black}{\fontsize{8pt}{1pt}\selectfont ODTSR}}
    \put(42.4,78.5){\color{black}{\fontsize{8pt}{1pt}\selectfont StrSR}}
    \put(50.9,78.5){\color{black}{\fontsize{8pt}{1pt}\selectfont FaithDiff}}
    \put(58.7,78.5){\color{black}{\fontsize{8pt}{1pt}\selectfont \textbf{HiFiPix (Ours)}}}
    \put(5.2,78.5){\color{black}{\fontsize{8pt}{1pt}\selectfont From RealLQ250}}
    
    \put(24.0,66.3){\color{black}{\fontsize{8pt}{1pt}\selectfont LQ}}
    \put(32.6,66.3){\color{black}{\fontsize{8pt}{1pt}\selectfont Dit4SR}}
    \put(41.9,66.3){\color{black}{\fontsize{8pt}{1pt}\selectfont LucidFlux}}
    \put(51.8,66.3){\color{black}{\fontsize{8pt}{1pt}\selectfont TSDSR}}
    \put(60.8,66.3){\color{black}{\fontsize{8pt}{1pt}\selectfont PiSA-SR}}
    \put(23.6,55.5){\color{black}{\fontsize{8pt}{1pt}\selectfont TVT}}
    \put(32.4,55.5){\color{black}{\fontsize{8pt}{1pt}\selectfont ODTSR}}
    \put(42.4,55.5){\color{black}{\fontsize{8pt}{1pt}\selectfont StrSR}}
    \put(50.9,55.5){\color{black}{\fontsize{8pt}{1pt}\selectfont FaithDiff}}
    \put(58.7,55.5){\color{black}{\fontsize{8pt}{1pt}\selectfont \textbf{HiFiPix (Ours)}}}
    \put(5.2,55.5){\color{black}{\fontsize{8pt}{1pt}\selectfont From RealLQ250}}
    
	\put(24.0,43.3){\color{black}{\fontsize{8pt}{1pt}\selectfont LQ}}
    \put(32.6,43.3){\color{black}{\fontsize{8pt}{1pt}\selectfont Dit4SR}}
    \put(41.9,43.3){\color{black}{\fontsize{8pt}{1pt}\selectfont LucidFlux}}
    \put(51.8,43.3){\color{black}{\fontsize{8pt}{1pt}\selectfont TSDSR}}
    \put(60.8,43.3){\color{black}{\fontsize{8pt}{1pt}\selectfont PiSA-SR}}
    \put(23.6,32.5){\color{black}{\fontsize{8pt}{1pt}\selectfont TVT}}
    \put(32.4,32.5){\color{black}{\fontsize{8pt}{1pt}\selectfont ODTSR}}
    \put(42.4,32.5){\color{black}{\fontsize{8pt}{1pt}\selectfont StrSR}}
    \put(50.9,32.5){\color{black}{\fontsize{8pt}{1pt}\selectfont FaithDiff}}
    \put(58.7,32.5){\color{black}{\fontsize{8pt}{1pt}\selectfont \textbf{HiFiPix (Ours)}}}
    \put(5.2,32.5){\color{black}{\fontsize{8pt}{1pt}\selectfont From RealLR200}}

    \put(0.8,16.2){\color{black}{\fontsize{8pt}{1pt}\selectfont LQ From RealLR200}}
    \put(18.5,16.2){\color{black}{\fontsize{8pt}{1pt}\selectfont Dit4SR}}
    \put(30.3,16.2){\color{black}{\fontsize{8pt}{1pt}\selectfont LucidFlux}}
    \put(45.1,16.2){\color{black}{\fontsize{8pt}{1pt}\selectfont TSDSR}}
    \put(58.8,16.2){\color{black}{\fontsize{8pt}{1pt}\selectfont PiSA-SR}}
    \put(5.5,1.1){\color{black}{\fontsize{8pt}{1pt}\selectfont TVT}}
    \put(18.0,1.1){\color{black}{\fontsize{8pt}{1pt}\selectfont ODTSR}}
    \put(32.4,1.1){\color{black}{\fontsize{8pt}{1pt}\selectfont StrSR}}
    \put(45.0,1.1){\color{black}{\fontsize{8pt}{1pt}\selectfont FaithDiff}}
    \put(56.5,1.1){\color{black}{\fontsize{8pt}{1pt}\selectfont \textbf{HiFiPix (Ours)}}}

	\end{overpic}
	\vspace{-1mm}
	\caption{Qualitative comparisons with existing methods on unpaired RealLQ250 and RealLR200 test sets.}
	\label{fig: Qualitative_supp2}
	\vspace{-3mm}
\end{figure*}

\begin{figure*}[t]
	\begin{overpic}[width=0.99\linewidth]{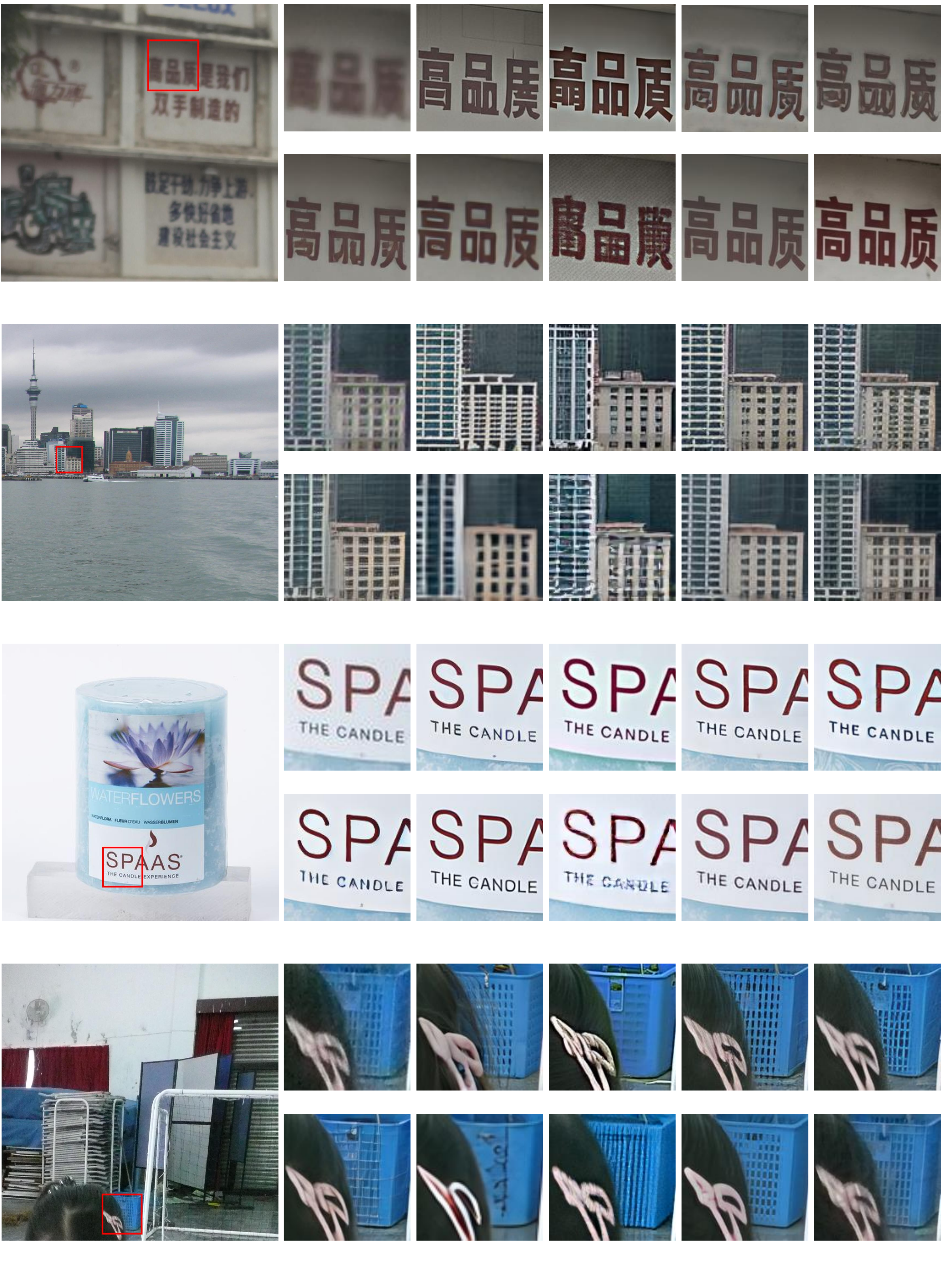}
    \put(26.0,88.6){\color{black}{\fontsize{8pt}{1pt}\selectfont LQ}}
    \put(35.0,88.6){\color{black}{\fontsize{8pt}{1pt}\selectfont Dit4SR}}
    \put(44.6,88.6){\color{black}{\fontsize{8pt}{1pt}\selectfont LucidFlux}}
    \put(55.1,88.6){\color{black}{\fontsize{8pt}{1pt}\selectfont TSDSR}}
    \put(65.4,88.6){\color{black}{\fontsize{8pt}{1pt}\selectfont PiSA-SR}}
    \put(25.5,77.0){\color{black}{\fontsize{8pt}{1pt}\selectfont TVT}}
    \put(35.0,77.0){\color{black}{\fontsize{8pt}{1pt}\selectfont ODTSR}}
    \put(45.4,77.0){\color{black}{\fontsize{8pt}{1pt}\selectfont StrSR}}
    \put(53.0,77.0){\color{black}{\fontsize{8pt}{1pt}\selectfont \textbf{HiFiPix (Ours)}}}
    \put(66.8,77.0){\color{black}{\fontsize{8pt}{1pt}\selectfont GT}}
    \put(3.5,77.0){\color{black}{\fontsize{8pt}{1pt}\selectfont From GGT-GeneralMix}}
    
    \put(26.0,63.7){\color{black}{\fontsize{8pt}{1pt}\selectfont LQ}}
    \put(35.0,63.7){\color{black}{\fontsize{8pt}{1pt}\selectfont Dit4SR}}
    \put(44.6,63.7){\color{black}{\fontsize{8pt}{1pt}\selectfont LucidFlux}}
    \put(55.1,63.7){\color{black}{\fontsize{8pt}{1pt}\selectfont TSDSR}}
    \put(65.4,63.7){\color{black}{\fontsize{8pt}{1pt}\selectfont PiSA-SR}}
    \put(25.5,52.0){\color{black}{\fontsize{8pt}{1pt}\selectfont TVT}}
    \put(35.0,52.0){\color{black}{\fontsize{8pt}{1pt}\selectfont ODTSR}}
    \put(45.4,52.0){\color{black}{\fontsize{8pt}{1pt}\selectfont StrSR}}
    \put(53.0,52.0){\color{black}{\fontsize{8pt}{1pt}\selectfont \textbf{HiFiPix (Ours)}}}
    \put(66.8,52.0){\color{black}{\fontsize{8pt}{1pt}\selectfont GT}}
    \put(3.5,52.0){\color{black}{\fontsize{8pt}{1pt}\selectfont From GGT-GeneralMix}}
    
	\put(26.0,38.9){\color{black}{\fontsize{8pt}{1pt}\selectfont LQ}}
    \put(35.0,38.9){\color{black}{\fontsize{8pt}{1pt}\selectfont Dit4SR}}
    \put(44.6,38.9){\color{black}{\fontsize{8pt}{1pt}\selectfont LucidFlux}}
    \put(55.1,38.9){\color{black}{\fontsize{8pt}{1pt}\selectfont TSDSR}}
    \put(65.4,38.9){\color{black}{\fontsize{8pt}{1pt}\selectfont PiSA-SR}}
    \put(25.5,27.2){\color{black}{\fontsize{8pt}{1pt}\selectfont TVT}}
    \put(35.0,27.2){\color{black}{\fontsize{8pt}{1pt}\selectfont ODTSR}}
    \put(45.4,27.2){\color{black}{\fontsize{8pt}{1pt}\selectfont StrSR}}
    \put(53.0,27.2){\color{black}{\fontsize{8pt}{1pt}\selectfont \textbf{HiFiPix (Ours)}}}
    \put(66.8,27.2){\color{black}{\fontsize{8pt}{1pt}\selectfont GT}}
    \put(3.5,27.2){\color{black}{\fontsize{8pt}{1pt}\selectfont From GGT-GeneralMix}}

    \put(26.0,14.0){\color{black}{\fontsize{8pt}{1pt}\selectfont LQ}}
    \put(35.0,14.0){\color{black}{\fontsize{8pt}{1pt}\selectfont Dit4SR}}
    \put(44.6,14.0){\color{black}{\fontsize{8pt}{1pt}\selectfont LucidFlux}}
    \put(55.1,14.0){\color{black}{\fontsize{8pt}{1pt}\selectfont TSDSR}}
    \put(65.4,14.0){\color{black}{\fontsize{8pt}{1pt}\selectfont PiSA-SR}}
    \put(25.5,2.5){\color{black}{\fontsize{8pt}{1pt}\selectfont TVT}}
    \put(35.0,2.5){\color{black}{\fontsize{8pt}{1pt}\selectfont ODTSR}}
    \put(45.4,2.5){\color{black}{\fontsize{8pt}{1pt}\selectfont StrSR}}
    \put(53.0,2.5){\color{black}{\fontsize{8pt}{1pt}\selectfont \textbf{HiFiPix (Ours)}}}
    \put(66.8,2.5){\color{black}{\fontsize{8pt}{1pt}\selectfont GT}}
    \put(3.5,2.5){\color{black}{\fontsize{8pt}{1pt}\selectfont From GGT-GeneralMix}}
    
	\end{overpic}
	\vspace{-1mm}
	\caption{Qualitative comparisons with existing Real-ISR methods on paired LQ-GT real-world test sets: GGT-GeneralMix.}
	\label{fig: Qualitative_supp3}
	\vspace{-3mm}
\end{figure*}

\begin{figure*}[t]
	\begin{overpic}[width=0.99\linewidth]{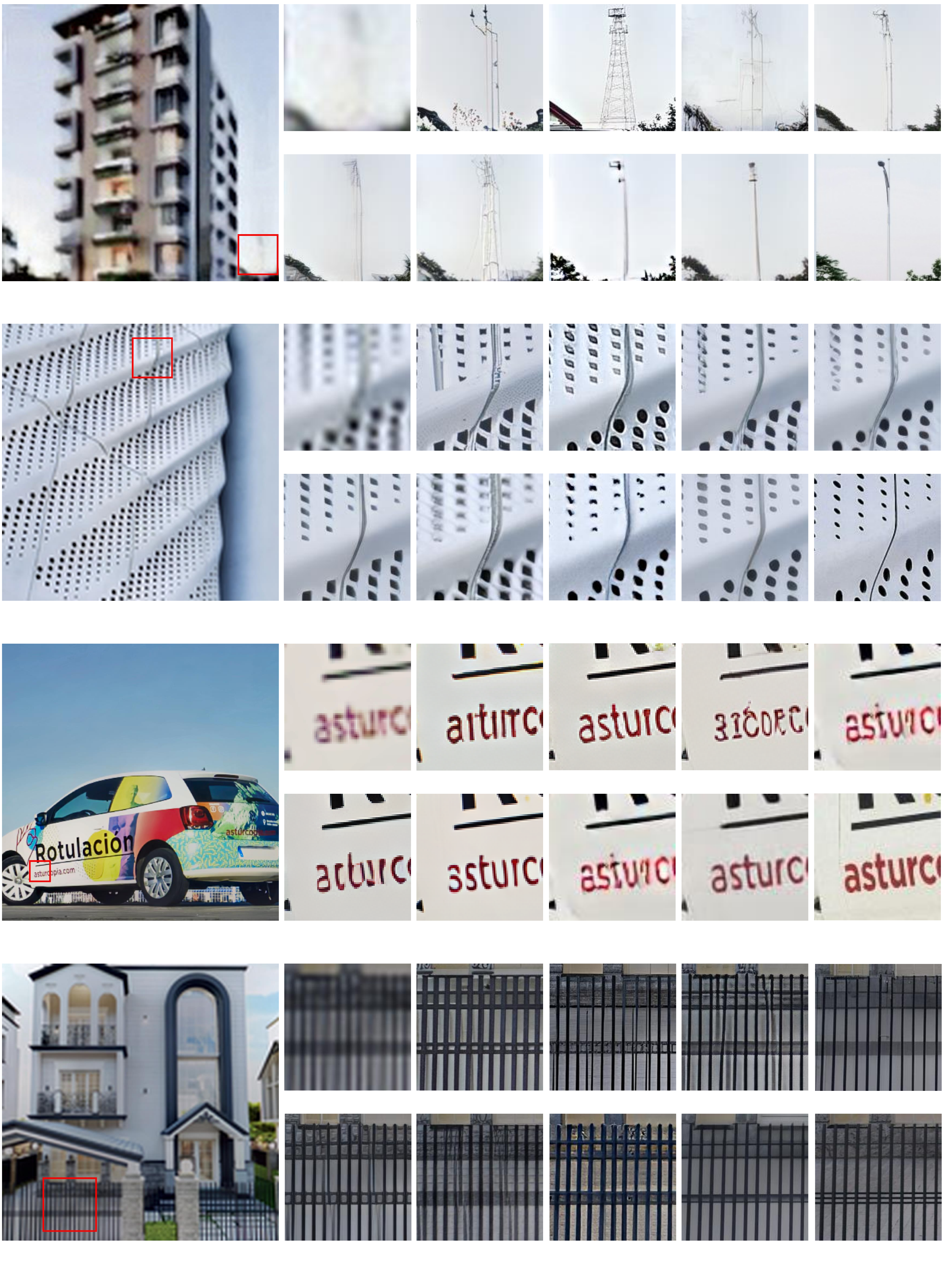}
    \put(26.0,88.6){\color{black}{\fontsize{8pt}{1pt}\selectfont LQ}}
    \put(44.6,88.6){\color{black}{\fontsize{8pt}{1pt}\selectfont LucidFlux}}
    \put(55.1,88.6){\color{black}{\fontsize{8pt}{1pt}\selectfont TSDSR}}
    \put(65.4,88.6){\color{black}{\fontsize{8pt}{1pt}\selectfont PiSA-SR}}
    \put(25.5,77.0){\color{black}{\fontsize{8pt}{1pt}\selectfont TVT}}
    \put(35.0,77.0){\color{black}{\fontsize{8pt}{1pt}\selectfont ODTSR}}
    \put(45.4,77.0){\color{black}{\fontsize{8pt}{1pt}\selectfont StrSR}}
    \put(53.0,77.0){\color{black}{\fontsize{8pt}{1pt}\selectfont \textbf{HiFiPix (Ours)}}}
    \put(66.8,77.0){\color{black}{\fontsize{8pt}{1pt}\selectfont GT}}
    \put(3.8,77.0){\color{black}{\fontsize{8pt}{1pt}\selectfont From our benchmark}}
    
    \put(26.0,63.7){\color{black}{\fontsize{8pt}{1pt}\selectfont LQ}}
    \put(35.0,63.7){\color{black}{\fontsize{8pt}{1pt}\selectfont Dit4SR}}
    \put(44.6,63.7){\color{black}{\fontsize{8pt}{1pt}\selectfont LucidFlux}}
    \put(55.1,63.7){\color{black}{\fontsize{8pt}{1pt}\selectfont TSDSR}}
    \put(65.4,63.7){\color{black}{\fontsize{8pt}{1pt}\selectfont PiSA-SR}}
    \put(25.5,52.0){\color{black}{\fontsize{8pt}{1pt}\selectfont TVT}}
    \put(35.0,52.0){\color{black}{\fontsize{8pt}{1pt}\selectfont ODTSR}}
    \put(45.4,52.0){\color{black}{\fontsize{8pt}{1pt}\selectfont StrSR}}
    \put(53.0,52.0){\color{black}{\fontsize{8pt}{1pt}\selectfont \textbf{HiFiPix (Ours)}}}
    \put(66.8,52.0){\color{black}{\fontsize{8pt}{1pt}\selectfont GT}}
    \put(3.8,52.0){\color{black}{\fontsize{8pt}{1pt}\selectfont From our benchmark}}
    
	\put(26.0,38.9){\color{black}{\fontsize{8pt}{1pt}\selectfont LQ}}
    \put(35.0,38.9){\color{black}{\fontsize{8pt}{1pt}\selectfont Dit4SR}}
    \put(44.6,38.9){\color{black}{\fontsize{8pt}{1pt}\selectfont LucidFlux}}
    \put(55.1,38.9){\color{black}{\fontsize{8pt}{1pt}\selectfont TSDSR}}
    \put(65.4,38.9){\color{black}{\fontsize{8pt}{1pt}\selectfont PiSA-SR}}
    \put(25.5,27.2){\color{black}{\fontsize{8pt}{1pt}\selectfont TVT}}
    \put(35.0,27.2){\color{black}{\fontsize{8pt}{1pt}\selectfont ODTSR}}
    \put(45.4,27.2){\color{black}{\fontsize{8pt}{1pt}\selectfont StrSR}}
    \put(53.0,27.2){\color{black}{\fontsize{8pt}{1pt}\selectfont \textbf{HiFiPix (Ours)}}}
    \put(66.8,27.2){\color{black}{\fontsize{8pt}{1pt}\selectfont GT}}
    \put(3.8,27.2){\color{black}{\fontsize{8pt}{1pt}\selectfont From our benchmark}}

    \put(26.0,14.0){\color{black}{\fontsize{8pt}{1pt}\selectfont LQ}}
    \put(35.0,14.0){\color{black}{\fontsize{8pt}{1pt}\selectfont Dit4SR}}
    \put(44.6,14.0){\color{black}{\fontsize{8pt}{1pt}\selectfont LucidFlux}}
    \put(55.1,14.0){\color{black}{\fontsize{8pt}{1pt}\selectfont TSDSR}}
    \put(65.4,14.0){\color{black}{\fontsize{8pt}{1pt}\selectfont PiSA-SR}}
    \put(25.5,2.5){\color{black}{\fontsize{8pt}{1pt}\selectfont TVT}}
    \put(35.0,2.5){\color{black}{\fontsize{8pt}{1pt}\selectfont ODTSR}}
    \put(45.4,2.5){\color{black}{\fontsize{8pt}{1pt}\selectfont StrSR}}
    \put(53.0,2.5){\color{black}{\fontsize{8pt}{1pt}\selectfont \textbf{HiFiPix (Ours)}}}
    \put(66.8,2.5){\color{black}{\fontsize{8pt}{1pt}\selectfont GT}}
    \put(3.8,2.5){\color{black}{\fontsize{8pt}{1pt}\selectfont From our benchmark}}
    
	\end{overpic}
	\vspace{-1mm}
	\caption{Qualitative comparisons with existing Real-ISR methods on our paired LQ-GT real-world benchmark: Dual-Q-Bench.}
	\label{fig: Qualitative_supp4}
	\vspace{-3mm}
\end{figure*}

\paragraph{Qualitative Comparisons.}
As shown in Fig.~\ref{fig: Qualitative_supp1}, Fig.~\ref{fig: Qualitative_supp2}, Fig.~\ref{fig: Qualitative_supp3}, and Fig.~\ref{fig: Qualitative_supp4}, HiFiPix achieves the best visual quality while preserving high-fidelity image details on various real-world test sets (\eg, RealSR, DrealSR, GGT-GeneralMix, RealLQ250, and our proposed Dual-Q-Bench), including legible text and natural textures. In contrast, VAE-based approaches tend to introduce twisted artifacts and over-smoothing details. Those visual comparisons
validate the superior performance of HiFiPix.
\subsection{Comparison of Model Parameters}

We compare generative backbone, language model, and trainable component parameters across existing Real-ISR methods in Tab.~\ref{tab: parameter}. Our generative backbone is substantially more compact than methods built on large generative architectures: ODTSR and LucidFlux use 22B Qwen and 12B FLUX-dev backbones, respectively, incurring heavy parameter overhead. Furthermore, removing the text encoder cuts our total parameters relative to SDXL-, SD3-, SD3.5-, and Z-Image-based approaches such as SUPIR, FaithDiff, DiT4SR, and StrSR. Our method requires merely 207.1M trainable parameters, surpassing all multi-step methods and delivering better parameter efficiency than single-step methods relying on large-scale generative backbones (e.g., ODTSR and StrSR). While SD2.1-based single-step methods have fewer trainable parameters, they yield poorer restoration performance. Together with our competitive restoration quality, these results verify our method’s superior efficiency in both model size and fine-tuning cost.

\subsection{Ablation Studies}
\label{sec:abl study in supp}

\textbf{Number of Selected Layers in LoRA-SP.} With $35$ LoRA layers total for fine-tuning, Tab.~\ref{tab: num_lora} examines how many layers should be pruned to boost generative capability while preserving fidelity. We find that both insufficient and excessive pruning yield suboptimal results. Specifically, pruning too few layers has only a limited effect on improving generative performance. While pruning more layers can improve generative performance, it severely compromises the model’s fidelity. Pruning $4$ layers of all LoRA layers achieves the optimal trade-off between fidelity and generative capability.

\textbf{Number of Distillation Layers in MixCSD.}
Tab.~\ref{tab: num_dis} analyzes the effect of intermediate layer quantity for MixCSD-based contrastive distillation. We find that insufficient distillation layers provide limited supervision, hindering the generation of high-frequency representations and thus impeding performance improvements. Conversely, distilling excessive layers imposes redundant constraints on intermediate features and reduces generative flexibility. Accordingly, we uniformly adopt $5$ layers (roughly $1/7$ of total layers) for contrastive distillation in final settings.

\textbf{Layer-Selective Freezing vs. Layer-Adaptive Attenuation.} We initially expected that layer-adaptive attenuation of LoRA across all layers would improve generative quality, as weakening the adapters should release the generative prior of the pretrained backbone. However, no gain was observed in practice: such attenuation only rescales the magnitude of each layer's update while preserving its direction, so the adaptation drift is weakened but never removed, and every layer is left in an intermediate state that matches neither the pretrained backbone nor the converged adapter. We therefore adopt a more decisive layer-selective freezing of LoRA instead.

\end{document}

%% file: math_commands.tex
\usepackage{amsmath,amsfonts,bm}

\def\eqref#1{equation~\ref{#1}}

\def\1{\bm{1}}

\DeclareMathAlphabet{\mathsfit}{\encodingdefault}{\sfdefault}{m}{sl}
\SetMathAlphabet{\mathsfit}{bold}{\encodingdefault}{\sfdefault}{bx}{n}

